\documentclass[pdflatex,sn-mathphys-num]{sn-jnl}
\usepackage{longtable}
\usepackage{graphicx}%
\usepackage{multirow}%
\usepackage{amsmath,amssymb,amsfonts}%
\usepackage{amsthm}%
\usepackage[title]{appendix}%
\usepackage{xcolor}%
\usepackage{textcomp}%
\usepackage{manyfoot}%
\usepackage{booktabs}%
\usepackage{algorithm}%
\usepackage{algorithmicx}%
\usepackage{algpseudocode}%
\usepackage{listings}%

\theoremstyle{thmstyleone}%
\theoremstyle{thmstyletwo}%
\theoremstyle{thmstylethree}%

\begin{document}

\title[Diffusion-Based Refinement for Precipitation Nowcasting]{Diffusion-Based Refinement for Kilometer-Scale Probabilistic Precipitation Nowcasting}


\author[1]{\fnm{Dohyun}~\sur{Park}}\email{do3204@snu.ac.kr}
\equalcont{These authors contributed equally to this work.}

\author[2]{\fnm{Changhoon}~\sur{Song}}\email{changhoon.song93@snu.ac.kr}
\equalcont{These authors contributed equally to this work.}

\author[2]{\fnm{Teng-Yuan}~\sur{Chang}}\email{ty.chang9408@gmail.com}

\author*[3]{\fnm{Yoo-Geun}~\sur{Ham}}\email{yoogeun@snu.ac.kr}

\author*[1]{\fnm{Youngjoon}~\sur{Hong}}\email{hongyj@snu.ac.kr}

\affil[1]{\orgdiv{Department of Mathematical Sciences}, \orgname{Seoul National University}, \orgaddress{\city{Seoul}, \country{Republic of Korea}}}
\affil[2]{\orgdiv{Research Institute of Mathematics}, \orgname{Seoul National University}, \orgaddress{\city{Seoul}, \country{Republic of Korea}}}
\affil[3]{\orgdiv{Department of Environmental Management}, \orgname{Seoul National University}, \orgaddress{\city{Seoul}, \country{Republic of Korea}}}

\abstract{Localized extreme precipitation is a major trigger of urban flash floods and landslides, yet producing nowcasts that combine fine spatial detail with probabilistic uncertainty remains challenging. Here we introduce exPreCast-ENS, a conditional residual diffusion framework that transforms the deterministic 4\,km radar nowcaster exPreCast into a 1\,km probabilistic ensemble while correcting systematic forecast errors. Conditioning on both the forecast and preceding radar observations lets the ensemble-mean correct the baseline rather than perturb it, while members represent unresolved fine-scale variability. Over the Korean Peninsula, skill improves with ensemble size. In two high-impact events in 2023, a 30-member ensemble recovers 38--47\% of heavy-rain pixels missed by exPreCast while retaining approximately 95\% of its correct detections and alarming on under 1\% of the pixels it correctly left clear. The method generates a 1-h forecast in 3.4\,s on a single GPU and yields consistent improvements on the French regional MeteoNet radar dataset.}

\maketitle


\section{Introduction}\label{sec:intro}
Accurate precipitation forecasting is essential for disaster preparedness and risk mitigation. This challenge is particularly pronounced in extreme precipitation nowcasting, as hazardous rainfall events often develop rapidly over localized regions. Effective warning systems therefore require forecasts with both high spatial resolution and frequent updates. However, conventional numerical weather prediction (NWP) struggles to meet these requirements because kilometer-scale precipitation forecasting demands substantial computational resources, particularly when uncertainty estimates are obtained through ensemble simulations.

The limitations of NWP have motivated research on data-driven weather prediction, especially using deep learning. Recently, forecasting systems such as Pangu-Weather~\cite{pangu-weather}, FourCastNet~\cite{fourcastnet}, GenCast~\cite{gencast}, and Aurora~\cite{aurora} have demonstrated that neural networks can achieve forecasting skill comparable to or even exceeding that of state-of-the-art NWP systems in a variety of atmospheric prediction tasks. These successes have, in turn, stimulated growing interest in high-resolution regional precipitation forecasting, where the computational burden of conventional approaches is particularly severe.

Radar-based approaches have attracted considerable attention because weather radar provides direct precipitation observations at high spatial and temporal resolutions. Representative examples include ConvLSTM~\cite{convlstm} and PredRNN~\cite{predrnn}, which model the spatiotemporal evolution of radar echoes using recurrent architectures, and Earthformer~\cite{earthformer}, which employs transformer architectures to capture long-range spatiotemporal dependencies. 
More recent precipitation-specific models have further sought to improve the representation of localized and intense rainfall. AlphaPre~\cite{alphapre}, for example, disentangles precipitation evolution into phase and amplitude components in the frequency domain to separately model changes in precipitation location and intensity. SimCast~\cite{simcast} employs short-to-long-term knowledge distillation together with a weighted objective that emphasizes heavy precipitation.
However, these deterministic models often produce overly smoothed forecasts and fail to represent localized extreme precipitation. Although several studies have introduced specialized objective functions, such as Fourier Amplitude Correlation Loss (FACL)~\cite{facl}, Wavelet-Fourier Composite Loss (WFCL)~\cite{wfcl}, and Adjusted Mean Squared Error (AMSE)~\cite{subich2025fixing}, to preserve fine-scale precipitation structures, deterministic models still provide only a single estimate of future precipitation and cannot represent the inherent uncertainty associated with precipitation evolution.

On the other hand, generative approaches yield probabilistic nowcasting, and can thereby represent multiple plausible future scenarios and quantify uncertainty. DGMR~\cite{dgmr} established this direction for radar nowcasting with a conditional generative adversarial network. More recent approaches use diffusion models. For instance, PreDiff~\cite{prediff} employs latent diffusion to model the distribution of future radar fields while allowing domain-specific knowledge to guide the generation process.
DiffCast~\cite{diffcast} and CasCast~\cite{cascast} have leveraged deterministic backbones, which first estimate the conditional mean of possible scenarios, and a diffusion model, which learns the residual distribution and generates probabilistic predictions. Compared with deterministic models, these diffusion-based approaches produce substantially sharper forecasts, demonstrating performance on benchmarks such as SEVIR from the contiguous United States~\cite{sevir} or MeteoNet from southeastern France~\cite{meteonet}, particularly in preserving fine-scale structures and intense precipitation. Furthermore, ensemble forecast results allow to reduce the forecast errors associated with the systematic uncertainties by taking averages of multiple forecast realizations. 
However, existing diffusion-based nowcasting frameworks typically require either iterative generation of the forecast field or additional latent-space encoding and decoding, which increases architectural and computational complexity. This motivates a lightweight probabilistic extension that directly refines an already skillful deterministic forecast, without relying on a separate latent representation.

In this study, we extend the deterministic exPreCast model~\cite{exPreCast} into exPreCast-ENS, a probabilistic forecasting framework that generates kilometer-scale predictions in real time.


Despite its deterministic mechanism, exPreCast is designed to preserve local precipitation patterns and generate forecasts as sharp as those produced by generative models, achieving state-of-the-art performance on KMA from South Korea~\cite{exPreCast}, SEVIR, and MeteoNet datasets.
In the proposed framework, we further enhance exPreCast with a lightweight diffusion module that enables the recovery of finer extreme rainfall signals missed by the deterministic forecast while barely introducing spurious precipitation.
This indicates that the diffusion module, which is introduced to complement exPreCast, generates reliable structure instead of randomized noise. Furthermore, multiple probabilistic samples can be leveraged to estimate uncertainty or to generate an ensemble forecast based on the consensus of samples. We also show that forecast skill improves consistently with ensemble size, demonstrating that the probabilistic extension captures a sufficiently diverse set of plausible precipitation evolutions. Despite these additional capabilities, the diffusion module contains relatively few parameters and produces one-hour forecast in seconds on a single GPU.

Our approach is related to CorrDiff~\cite{corrdiff}, which combines regression and diffusion model. The regression model first estimates a conditional mean field from a low-resolution input, and the subsequent diffusion model generates high-resolution residual realizations around the field. This formulation has proven effective for static downscaling and motivated us to adopt residual diffusion for forecasting systems. However, forecasting relying on spatial refinement is challenging since it involves systematic errors that evolve with lead time. To address this challenge, we combine historical radar observations and deterministic forecast. The historical context enables the diffusion model to capture forecast errors of the deterministic backbone based on the real observation. As a result, the residual distribution corrects systematic errors while simultaneously generating probabilistic forecasts.
We verify this behavior empirically by averaging multiple diffusion samples. If the diffusion module only generated stochastic perturbations with zero expectation, the average over samples would remain close to the deterministic forecast. Instead, we will demonstrate that the distribution has a non-zero mean and hence corrects biases in exPreCast results. This distinguishes the proposed framework from static residual downscaling and suggests that residual diffusion can serve as a general method for correcting the systematic errors of the deterministic nowcasting models. In addition, the framework can be adopted without explicit downscaling. Experiments on MeteoNet will show that the proposed framework is still effective without explicit downscaling. While the KMA experiments focus on recovering 1 km precipitation structures from deterministic 4 km forecasts, experiments on the MeteoNet dataset show that the same framework benefits even when the residual has the same spatial resolution as the inputs to deterministic backbone. 

The main contributions of this study are as follows. First, we propose a lightweight probabilistic precipitation nowcasting framework that combines the exPreCast with a residual diffusion module. The proposed framework generates high-resolution probabilistic forecasts in real-time while requiring only a modest number of parameters and seconds of inference on a single GPU. Second, we demonstrate that residual diffusion can recover localized extreme precipitation structures missed by deterministic backbone. Furthermore, forecast skill improves consistently with ensemble size, indicating that the generated samples capture a diverse set of plausible precipitation evolutions and provide meaningful ensemble guidance. 
Third, we show that the learned residual distribution is not merely zero-mean stochastic variability around the deterministic forecast. Its ensemble-mean remains systematically nonzero and corrects biases in the deterministic backbone, rather than merely introducing random perturbations.
Lastly, we validate the proposed framework on both the KMA and MeteoNet datasets. While the KMA experiments demonstrate the ability to recover missing structures in coarse observations, the MeteoNet experiments show that the same strategy remains beneficial even without explicit downscaling.

\begin{figure}[!htbp]
\centering
\includegraphics[width=\textwidth]{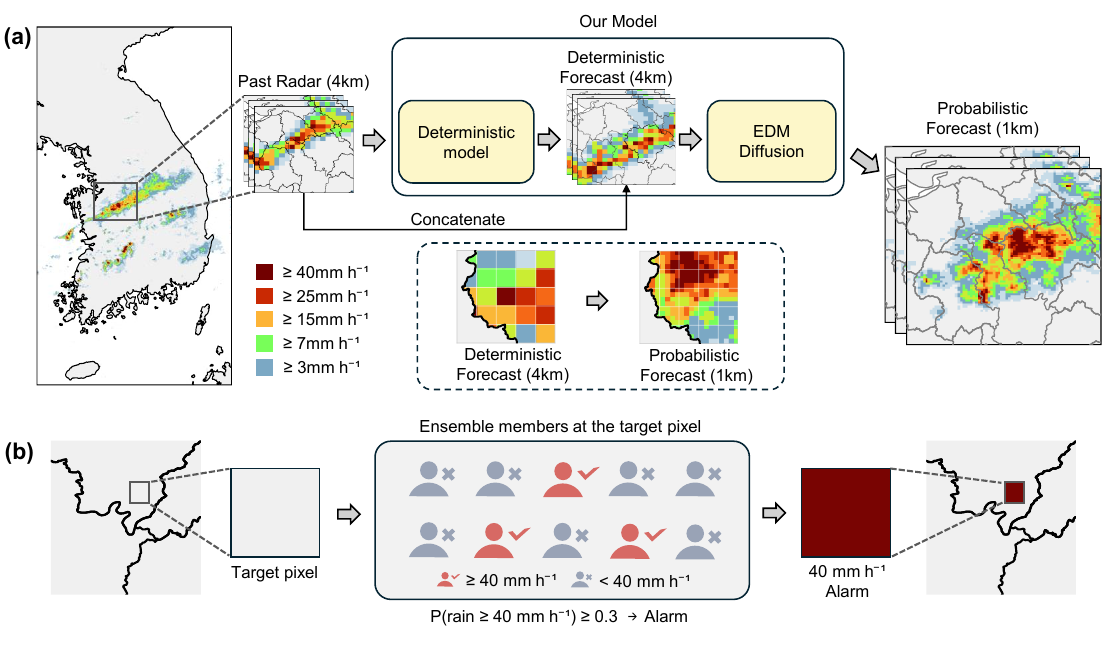}
\caption{\textbf{exPreCast-ENS: residual diffusion extension of exPreCast for high-resolution probabilistic nowcasting.} (a)~The deterministic backbone (exPreCast) takes a 4\,km past radar sequence and produces a 4\,km deterministic forecast; an EDM-based diffusion module, conditioned jointly on this deterministic forecast and on the same past radar sequence, then produces a 1\,km probabilistic forecast. (b)~Alarm probabilities are derived from the fraction of ensemble members exceeding a given intensity threshold at each pixel. Each icon is one member, red where it exceeds the threshold and grey where it does not; in the example shown, 3 of 10 members exceed $40\,\text{mm\,h}^{-1}$, so $P=0.3$ and the pixel meets the $P\geq 0.3$ criterion. The selection rationale for this criterion is provided in Supplementary Fig. 1.}\label{fig:overview}
\end{figure}


\section{Results}
\label{sec:results}
We produce a probabilistic precipitation forecast on a high-resolution grid via exPreCast-ENS, a two-stage framework: a deterministic backbone first forecasts the precipitation field on a coarse grid, which captures the overall structure of the rainfall, and then a residual diffusion module transforms this into a finer forecast. We adopt exPreCast~\cite{exPreCast}, which achieved the state-of-the-art performance on various radar datasets, as the deterministic backbone and EDM~\cite{edm} training for the diffusion module. 
The backbone serves as a strong deterministic baseline, and we investigate whether the proposed diffusion refinement provides benefits beyond coarse-grid forecasting in three aspects: forecast skill, distributional fidelity, and pattern recovery.

We train and evaluate the proposed framework on two radar precipitation datasets, KMA~\cite{exPreCast} and MeteoNet~\cite{meteonet}. Since KMA spans a relatively balanced spectrum of precipitation intensities, from ordinary rainfall to extreme events, we mainly focus on this dataset.
In KMA, the deterministic backbone operates on a 4\,km grid and our model refines the forecast to a 1\,km grid. This allows us to clearly assess the benefit of high-resolution probabilistic refinement over coarse-grid deterministic forecasting. We further validate our framework on the MeteoNet dataset, a widely used benchmark in which the deterministic backbone already generates prediction on 1\,km grid. The evaluation on MeteoNet further examines whether the proposed framework can provide improvements through probabilistic forecasting alone, without explicit super-resolution.

\begin{figure}[!htbp] 
\centering
\includegraphics[width=\textwidth]{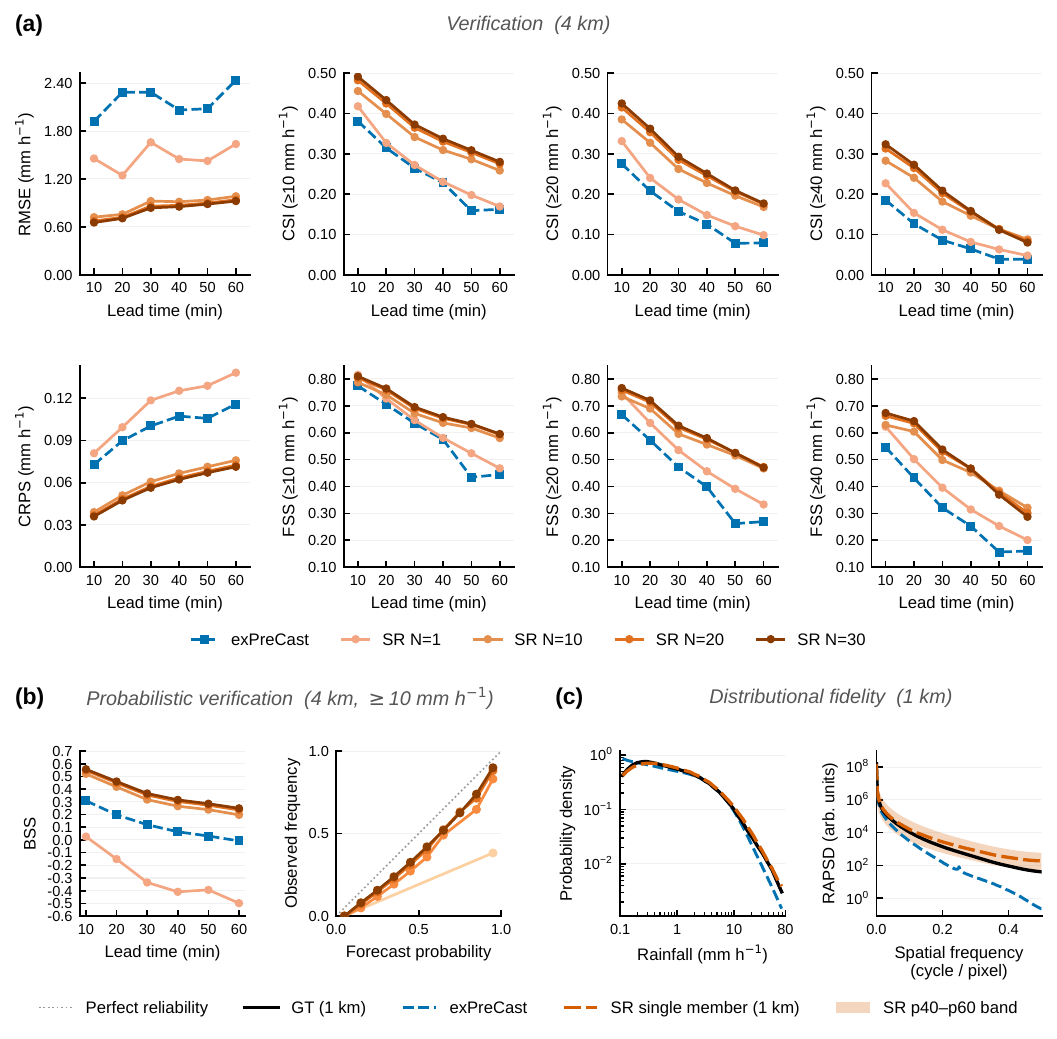}
\caption{\textbf{Quantitative evaluation on the KMA dataset.}
The backbone operates at 4\,km, and the diffusion module produces a 1\,km super-resolution (SR) forecast; all panels compare the exPreCast-ENS ensemble, labeled SR, against the exPreCast baseline.
(a)~Skill on the 4\,km backbone grid as a function of lead time, with one curve per ensemble size $N\in\{1,10,20,30\}$: RMSE and CRPS; CSI and FSS at the 10, 20, and 40\,$\text{mm\,h}^{-1}$ thresholds. The SR members are pooled to the 4\,km grid before scoring, and FSS uses a $3\times3$ neighborhood. A pixel counts as an alarm where the fraction of members exceeding the threshold is at least $0.3$; the choice of this value is explained in Supplementary Fig. 1. For the deterministic baseline, which is a single-member forecast, CRPS reduces to its MAE.
(b)~Probabilistic skill at the $10\,\text{mm\,h}^{-1}$ threshold: BSS as a function of lead time, with one curve per ensemble size, and the reliability diagram.
(c)~Distributional properties on the 1\,km target grid: rainfall PDF and RAPSD, showing the ground truth (solid black), the bilinearly interpolated exPreCast forecast (dashed blue), and one individual diffusion member (dashed orange). The member curve is computed per scene and lead time; the line is the median over all test scenes and the shaded band the 40th--60th percentile across scenes, averaged over lead times. Ground truth and exPreCast are deterministic and shown as the median only.}\label{fig:kma-eval}
\end{figure}

\subsection{Skill and distribution on KMA}\label{sec:results-ensemble}
We first present quantitative and qualitative evaluations of the proposed framework on the KMA dataset. We analyze deterministic and probabilistic forecast skills, evaluate distributional fidelity, and visually confirm that our result recovers fine-scale precipitation structures that are not resolved in coarse-grid predictions.

\paragraph*{Deterministic and probabilistic forecast skill with ensemble size.}
For a fair comparison with the deterministic exPreCast baseline, we evaluate both forecasts at the baseline's native 4\,km resolution. We use the official exPreCast weights and upscale our 1\,km probabilistic forecast to the coarser 4\,km grid, so that the comparison reflects forecasting skill rather than differences in spatial resolution.
Furthermore, we generate $N$ samples, aggregate the upscaled samples, and report scores for different ensemble sizes, $N=1,10,20,30$. The ensemble method depends on the type of scores: Root Mean Square Error (RMSE) is computed between the average of samples and the ground truth; 
Critical Success Index (CSI) and Fractions Skill Score (FSS) are computed after converting the samples into a binary forecast according to \textit{alarm mask} (see Section~\ref{sec:methods}); Brier Skill Score (BSS), the reliability diagram, and Continuous Ranked Probability Score (CRPS) are computed directly from the empirical distribution formed by the $N$ samples.

Figure~\ref{fig:kma-eval}(a) shows the metrics at each lead time and demonstrates that performance improves consistently with ensemble size. The substantial reduction in RMSE suggests that the diffusion module corrects systematic biases, thereby allowing the forecast to better capture overall precipitation intensity. 
The lower CRPS indicates improved probabilistic forecast performance. 
The higher CSI and FSS indicate that the proposed model more accurately predicts rainfall events across multiple intensity thresholds while better preserving their spatial structure. The BSS in Figure~\ref{fig:kma-eval}(b) increases with ensemble size at every lead time and is positive for $N\geq 10$, showing that the predicted rain probabilities become more informative as members are added. The reliability diagram in the same panel shows the curve moving toward the diagonal as members are added, so the exceedance probabilities become better calibrated with ensemble size. Exact values are given in Supplementary Table 1 for the 4\,km verification grid and Supplementary Table 2 for the 1\,km grid.

\paragraph{Distributional fidelity at the finer resolution.}
To assess the distributional and spatial fidelity of the generated forecasts at the target 1\,km resolution, we use one independently generated member per test case and compare its rainfall probability density function (PDF) and radially averaged power spectral density (RAPSD) with those of the radar observations. Figure~\ref{fig:kma-eval}(c) shows that the individual super-resolution (SR) forecasts more closely reproduce both the observed rainfall-intensity distribution and the spatial power spectrum than the deterministic baseline bilinearly interpolated to 1\,km. These results indicate that the diffusion module restores realistic fine-scale precipitation variability beyond the spatial scales represented by the coarse deterministic forecast.

\begin{figure}[!htbp]
\centering
\includegraphics[width=\textwidth]{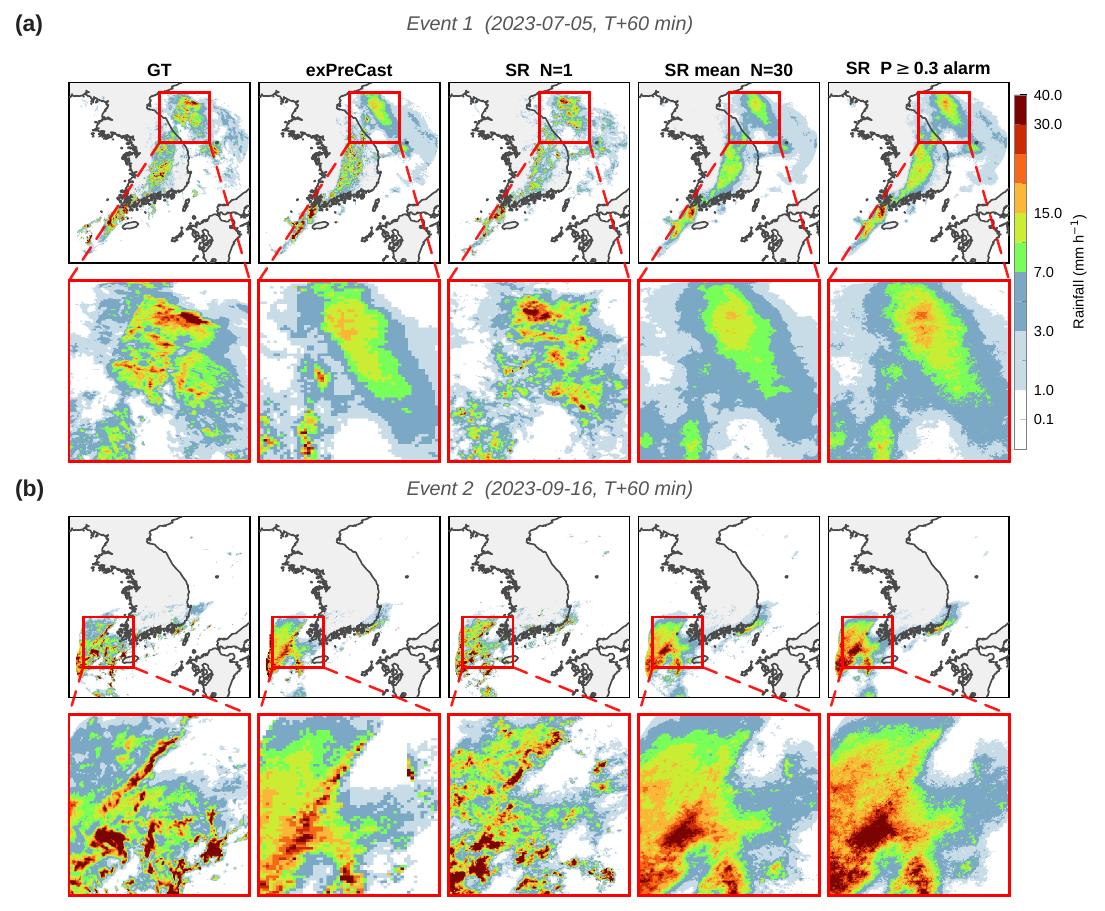}
\caption{\textbf{Three SR products compared with the backbone on two KMA events.} Each row compares, from left to right, the ground truth (GT), the exPreCast backbone, a single SR member, the 30-member SR mean, and the alarm mask, for (a)~2023-07-05 and (b)~2023-09-16, both at $T+60$\,min. The backbone is at 4\,km and the SR outputs at 1\,km. The alarm panel is not a binary mask: each pixel is colored by the highest level of the shared rainfall scale that at least $30\%$ of the members exceed, so it can be read against the other columns directly. Thresholding this panel at a given level recovers the binary alarm mask $M_{\mathrm{alarm}}$ of Section~\ref{sec:methods-diffusion} at that level. In both events, the individual SR member retains fine-scale structures that are smoothed out by the deterministic backbone, while the alarm mask expands the predicted heavy-rain region where the backbone underpredicts rainfall and excludes areas where it overpredicts rainfall.}\label{fig:sr-compare}
\end{figure}

\paragraph{Visual characteristics of the forecast products.}
Quantitative results indicated that the proposed framework improves both forecast skill and distributional fidelity, while Figure~\ref{fig:sr-compare} illustrates the complementary information provided by different uses of the generated samples. Compared with the deterministic backbone, an individual SR member recovers finer-scale textures and more localized extremes along the central rainband, representing one plausible realization of the future precipitation field. The mean of 30 SR members provides a spatially stable estimate of the expected precipitation field, but averaging attenuates sharp and localized extremes. By contrast, the alarm mask aggregates threshold-exceedance events across the ensemble rather than averaging rainfall intensities. It therefore highlights localized heavy-rain signals that may be diluted in the ensemble-mean and identifies regions in which heavy rainfall is supported by a substantial fraction of the ensemble members.

\subsection{Non-zero mean correction and spatial redistribution}\label{sec:results-redist}

\begin{figure}[!htbp]
\centering
\includegraphics[width=\textwidth]{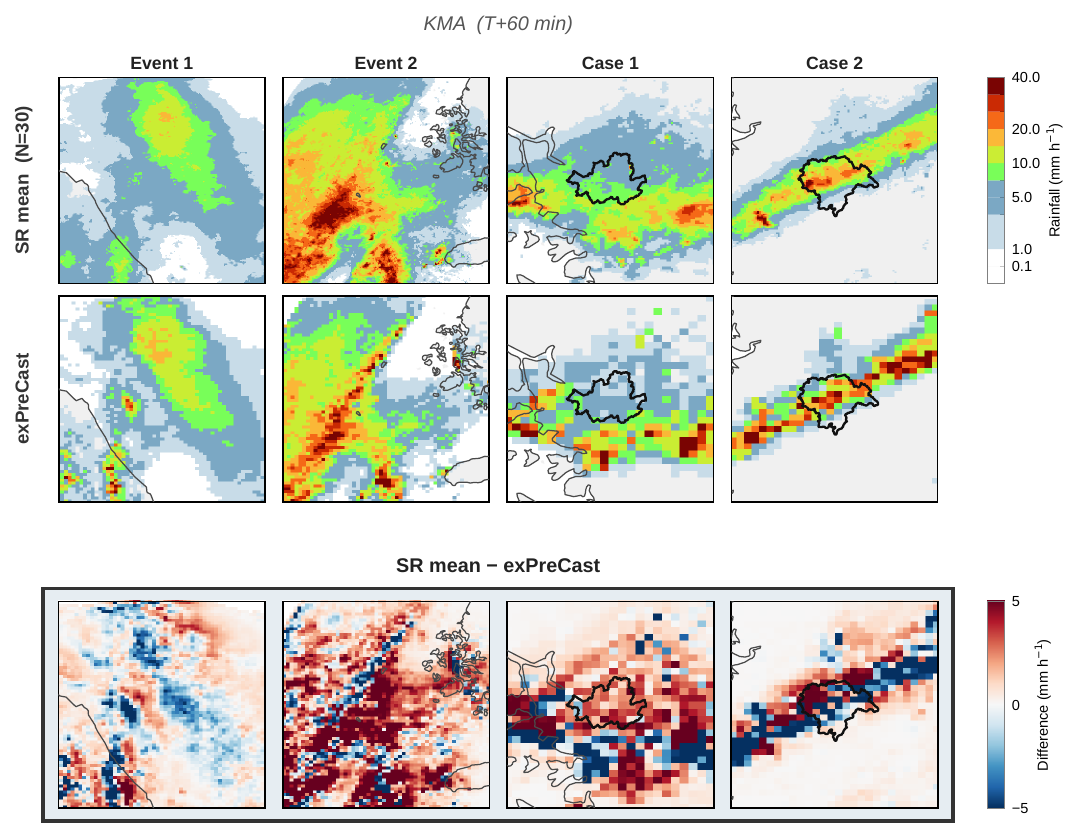}
\caption{\textbf{The 30-member mean against the backbone, across four KMA events.}
Top to bottom: the 30-member SR mean, the exPreCast backbone forecast, and the difference
between them, at $T+60$\,min. Columns are the two events of Fig.~\ref{fig:sr-compare} and
the two city cases of Fig.~\ref{fig:cases} (Seoul; Cheongju, with the city boundary
outlined). The SR mean is at 1\,km and exPreCast at its native 4\,km. The difference is
computed on the 4\,km grid, with the SR mean mean-pooled and the backbone forecast
subtracted from it, following the verification convention of
Section~\ref{sec:methods-eval}. The top two rows share the rainfall colormap; the bottom
row uses a diverging colormap centered at zero, red where the ensemble adds rain to the
backbone and blue where it removes it, saturating at $\pm 5\,\text{mm\,h}^{-1}$.}
\label{fig:residual}
\end{figure}

The ensemble-mean remains distinct from the deterministic backbone, indicating that the learned residual distribution has a non-zero conditional mean. If each probabilistic forecast consisted of the deterministic prediction plus a zero-mean stochastic residual, averaging an increasing number of ensemble members would suppress the stochastic component and recover the deterministic forecast. By contrast, although the ensemble-size results have largely stabilized by 30 members, the 30-member mean remains markedly different from the backbone prediction, as shown in Figure~\ref{fig:residual}. The difference becomes smoother as members are added but remains spatially structured and non-zero (Supplementary Fig. 2a,b).

The top and middle rows compare the 30-member SR mean with the deterministic exPreCast forecast. The SR mean exhibits more spatially coherent precipitation regions and finer intensity variations, whereas the backbone retains the coarse, block-like structures associated with its native 4\,km resolution. Because the two forecasts have different native resolutions, the bottom row isolates the ensemble-mean correction on a common 4\,km grid by mean-pooling the SR mean before subtracting the backbone forecast. The resulting difference fields contain substantial positive and negative corrections. Positive values indicate locations where the diffusion module increases the predicted rainfall, whereas negative values indicate locations where it reduces the backbone prediction. In the final two cases, the sign-changing corrections along the rainband further show that the ensemble-mean modifies both the intensity and spatial placement of the predicted precipitation.

These corrections primarily redistribute precipitation spatially rather than substantially altering its domain-integrated amount. Over the full 2023 test period (Table~\ref{tab:residual-fraction}), the net difference in domain-integrated precipitation between the backbone and the ensemble-mean is approximately $4\%$ of the backbone total, whereas their absolute difference is approximately $40\%$ of the same total. The former retains the sign of the local changes, so increases and decreases cancel, whereas the latter accumulates their magnitude. The large separation between these quantities indicates that substantial local changes largely cancel at the domain level: precipitation is increased in some regions and reduced in others while the domain-integrated amount remains nearly unchanged. The same comparison restricted to the observed rain area gives a smaller net difference and a larger absolute difference, so the correction is concentrated where precipitation occurs. Together with the improved verification scores in Figure~\ref{fig:kma-eval} and the comparisons with radar observations in Figure~\ref{fig:sr-compare}, these results show that the diffusion module learns a structured, non-zero-mean correction to the deterministic field rather than simply superimposing random fine-scale variability.

\begin{table}[!htbp]
\small
\caption{\textbf{The correction as a fraction of the backbone rainfall (KMA 4\,km grid, full 2023 test period, $N=30$).} The difference between the 30-member SR mean and the exPreCast forecast, expressed as a fraction of the backbone rainfall over the same pixels. The net term sums the difference with sign, so local increases and decreases cancel; the absolute term sums its magnitude regardless of sign. Their ratio measures how far the local changes exceed the net change. Values resolved by lead time are given in Supplementary Table 3.}\label{tab:residual-fraction}
\begin{tabular*}{\textwidth}{@{\extracolsep\fill}lccc}
\toprule
Evaluation set & Net (\%) & Absolute (\%) & Ratio \\
\midrule
Full domain & $+3.90$ & $39.53$ & $10.1$ \\
Observed rain area ($\text{GT}\geq 0.1\,\text{mm\,h}^{-1}$) & $+2.05$ & $45.49$ & $22.2$ \\
\bottomrule
\end{tabular*}
\end{table}

\subsection{Event-level verification through case studies}\label{sec:results-cases}
\begin{figure}[!htbp]
\centering
\includegraphics[width=\textwidth]{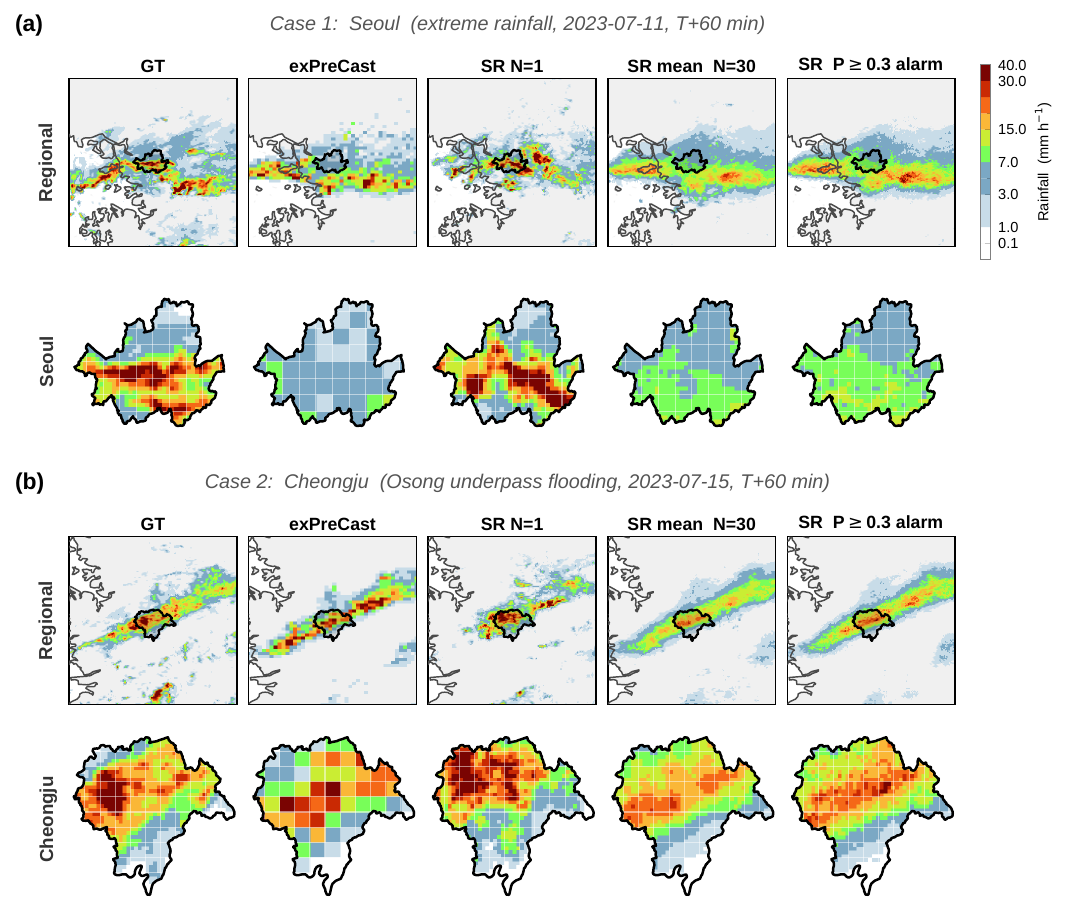}
\caption{\textbf{City-scale case studies.} (a)~Case 1, Seoul extreme rainfall (2023-07-11, $T+60$\,min); (b)~Case 2, Cheongju underpass flooding (2023-07-15, $T+60$\,min). For each case the top row is the regional view and the bottom row zooms to the city, with the city boundary outlined. Each row compares the ground truth, the exPreCast backbone, a single SR member, the 30-member SR mean, and the alarm mask. In (a) the single member shown is one that places the rainband over central Seoul; members differ in the position of the band. The SR mean and the alarm panel use all 30 members. The backbone is at 4\,km and the SR outputs at 1\,km. The alarm panel is not a binary mask: each pixel is colored by the highest level of the shared rainfall scale that at least $30\%$ of the members exceed, so it can be read against the other columns directly. Thresholding this panel at a given level recovers the binary alarm mask $M_{\mathrm{alarm}}$ of Section~\ref{sec:methods-diffusion} at that level.}\label{fig:cases}
\end{figure}

The preceding analyses show that the diffusion module restores fine-scale precipitation variability at 1\,km resolution and spatially redistributes the coarse deterministic forecast through a structured, non-zero-mean correction. We next examine how these properties translate to individual high-impact rainfall events. Figure~\ref{fig:cases} compares the radar observations with the deterministic backbone forecast, an individual SR member, the mean of 30 SR members, and the probability-thresholded alarm product.

\paragraph{Case 1: Extreme rainfall in Seoul.}
The first case corresponds to the extreme rainfall event over Seoul on 11 July 2023. As shown in Figure~\ref{fig:cases}(a), the ground truth contains a strong band crossing the central Seoul. However, the backbone places it too far south, missing most of the heavy-rain regions in the city. The ensemble, however, contains members that move the band northward and place hazardous rainfall over the downtown area; the member shown in Figure~\ref{fig:cases}(a) is one of these. Combining 30 individual samples, the ensemble-mean mitigates positional uncertainty in each sample, while the alarm mask recovers the rain across the city most sharply.

\paragraph{Case 2: Heavy rainfall associated with the flooding in Cheongju.}
The second case examines the heavy-rain event near Cheongju on 15 July 2023, which was associated with the Osong underpass flooding. Figure~\ref{fig:cases}(b) shows that the backbone well captures the location and intensity of the rain band, but the rainfall structures are missing due to the coarse grid. This makes the heaviest rainfall within the city elusive. A single prediction resolves the rain band onto the finer grid, reproduces it with intensity varying inside every pixel on the coarser grid, and reveals localized extreme rains relevant to the flood. The ensemble-mean preserves the overall rainband while reducing spatial uncertainty, resulting in a smoother but more consistent representation. The alarm mask further highlights where hazardous rainfall is likely to occur across ensemble members, making it more explicit than in the deterministic backbone or the ensemble-mean alone.

\begin{figure}[!htbp]
\centering
\includegraphics[width=\textwidth]{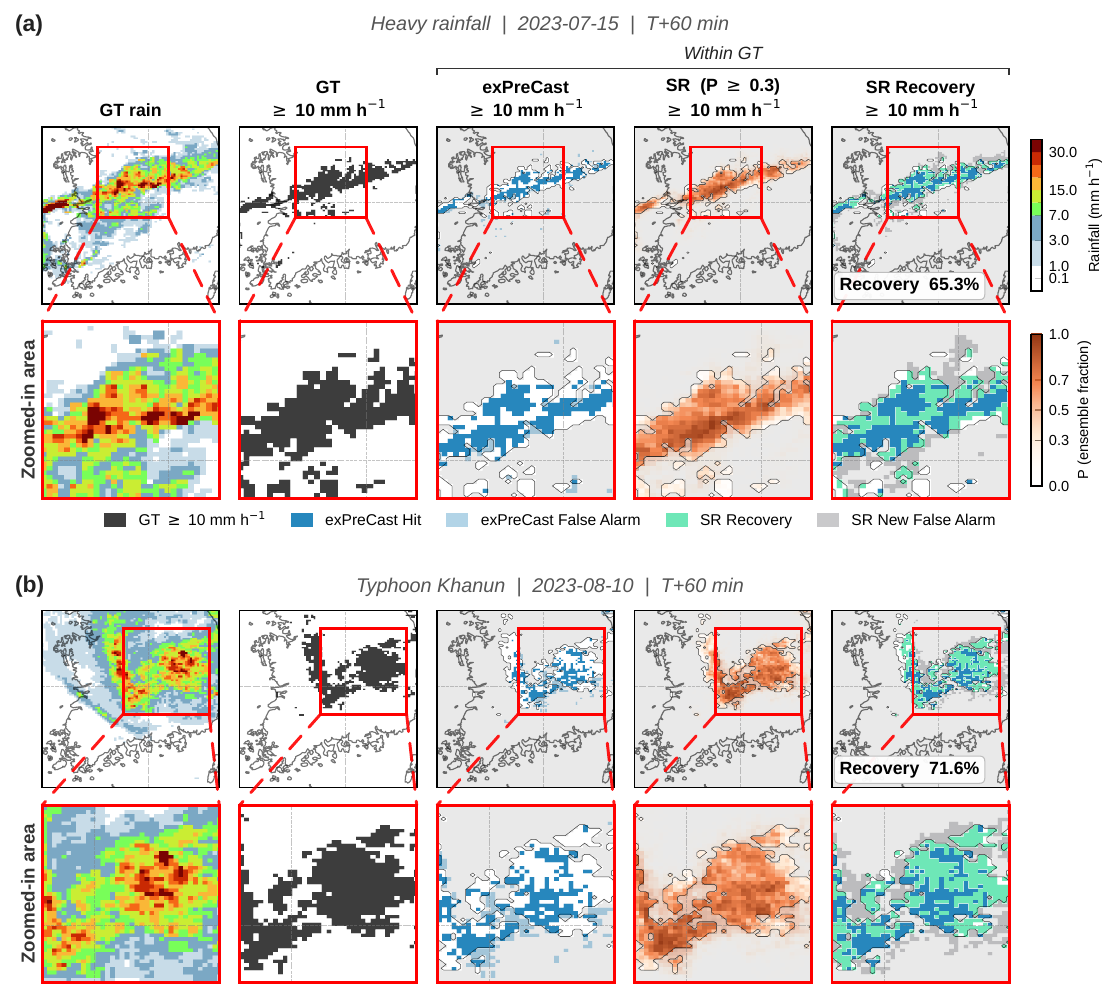}
\caption{\textbf{Recovery maps.} (a)~Monsoon event (initialized 2023-07-15 02:10 KST); (b)~Typhoon Khanun (initialized 2023-08-09 23:10 KST). Both are $+60$\,min forecasts on the 4\,km verification grid, at one time step within the periods aggregated in Table~\ref{tab:recovery}. Each row shows, from left to right, the GT rainfall, the GT $\geq 10\,\text{mm\,h}^{-1}$ mask, the exPreCast detection, the SR alarm at $P\geq 0.3$ (colored by the fraction of members exceeding $10\,\text{mm\,h}^{-1}$), and a combined recovery map; the lower panel of each row zooms to the boxed region. The three detection panels share the reference frame marked \textit{Within GT}: the contour is the GT $\geq 10\,\text{mm\,h}^{-1}$ boundary and the shading covers the observed rain area. Alarms are drawn wherever they fall, including outside that area, so alarms raised over dry ground remain visible. The recovery map separates five mutually exclusive categories, following the legend: GT $\geq 10\,\text{mm\,h}^{-1}$; exPreCast hits; exPreCast false alarms; SR recoveries (GT $\geq 10$ missed by exPreCast and recovered by SR); and SR new false alarms (SR alarms where GT $< 10$ and exPreCast did not alarm). The recovery rate printed on each map is the value for the time step shown; the corresponding rates aggregated over each 48-hour event window are given in Table~\ref{tab:recovery}, and the conditional rates are defined in Section~\ref{sec:methods-eval}.}\label{fig:recovery}
\end{figure}

\paragraph{Conditional recovery relative to the backbone.}
To quantify this behavior, we further evaluate the extent to which the SR alarm recovers observed heavy-rain pixels missed by the deterministic backbone. Figure~\ref{fig:recovery}(a) decomposes the predictions at an intensity threshold of $10\,\mathrm{mm}\,\mathrm{h}^{-1}$. The first two columns show the observed rainfall field and its corresponding heavy-rain mask, while the next two columns show the exPreCast detections and the ensemble exceedance probabilities, respectively. The final column combines these results into a conditional recovery map. Green pixels denote observed heavy-rain locations missed by exPreCast but recovered by the SR alarm. Gray pixels denote new SR false alarms, namely locations at which neither the observation nor exPreCast exceeds the prescribed threshold but the SR alarm is activated. The recovered pixels are concentrated primarily along the observed rainband rather than being scattered across unrelated parts of the domain. Although some new false alarms occur near the margins of the precipitation system, they remain largely adjacent to the observed heavy-rain area.

We quantify these outcomes by conditioning the evaluation on both the observation and the backbone prediction. In particular, the recovery rate measures the fraction of observed heavy-rain pixels missed by exPreCast that are subsequently detected by the SR alarm. The new false-alarm rate measures how frequently the SR alarm is activated at pixels that are dry in the observation and are not detected by exPreCast. For pixels already detected by the backbone, the hit-retention rate and hit-loss rate measure whether correct detections are preserved, and the false-alarm removal rate measures how often existing false alarms are deleted. This conditional decomposition distinguishes useful recovery from newly introduced false alarms and reveals changes relative to the deterministic backbone that are not resolved by aggregate scores such as CSI or probability of detection (POD) alone.

The results show that the SR alarm recovers a substantial fraction of the heavy-rain pixels missed by exPreCast, while introducing relatively few new false alarms, most of which lie close to the observed rainband. Among pixels already detected by exPreCast, the ensemble retains most correct detections while preferentially removing false alarms. Recovering a missed event requires positive residual realizations sufficiently large to cross the intensity threshold, whereas removing a backbone detection requires most ensemble members to fall below that threshold. The latter behavior cannot be produced by a strictly additive refinement that only increases precipitation. These conditional changes are therefore consistent with the bidirectional residual corrections described in Section~\ref{sec:results-redist}. Table~\ref{tab:recovery} summarizes the conditional statistics, together with CSI and POD, for the 15 July monsoon event and Typhoon Khanun.

\paragraph{Case 3: Extreme rainfall associated with Typhoon Khanun.}
We next examine an extreme rainfall event associated with Typhoon Khanun on 10 August 2023 to determine whether the recovery behavior observed in the monsoon event extends to a broader and more spatially extensive precipitation system. As shown in Figure~\ref{fig:recovery}(b), the deterministic backbone detects part of the observed heavy-rain area but leaves a substantial portion undetected. The probability-thresholded SR alarm recovers many of these missed heavy-rain pixels. Importantly, the recovered pixels are concentrated within or along the margins of the observed precipitation system rather than being scattered across unrelated regions. The conditional statistics in Table~\ref{tab:recovery} corroborate this visual result, showing substantial recovery of the heavy-rain pixels missed by exPreCast while the number of newly introduced false alarms remains limited.

Taken together, these case studies illustrate how the statistical improvements identified above translate to individual high-impact events. The proposed framework does not simply expand the predicted heavy-rain area indiscriminately. Instead, it recovers missed detections in spatially coherent regions, retains most of the correct detections made by the deterministic backbone, and introduces relatively few new false alarms.

\begin{table}[!htbp]
\small
\caption{\textbf{Event-level conditional verification, $\geq 10\,\text{mm\,h}^{-1}$ at $+60$\,min.} exPreCast (Base; the 4\,km baseline) against the 30-member SR ensemble (SR) at the $P\geq 0.3$ alarm criterion, on the 4\,km verification grid, for two high-impact events of 2023. The five conditional rates split the correction by what the baseline did at each pixel. Where the baseline misses ($\text{Base}=0$), SR either recovers heavy rain that fell or alarms where none did. Where the baseline fires ($\text{Base}=1$), SR keeps a correct detection, deletes one, or deletes a false one. $N=30$; 288 forecasts per event, at a 10-min cadence. Corresponding rates over the full 2023 test period, across all lead times and thresholds, are given in Supplementary Table 4.}\label{tab:recovery}
\begin{tabular*}{\textwidth}{@{\extracolsep\fill}l l cc}
\toprule
& & Monsoon & Typhoon Khanun \\
& & (2023-07-14/15) & (2023-08-09/10) \\
\midrule
\multirow{2}{*}{CSI}    & Base & 0.210 & 0.157 \\
                        & SR & \textbf{0.351} ($+67\%$) & \textbf{0.295} ($+89\%$) \\
\midrule
\multirow{2}{*}{POD}    & Base & 0.251 & 0.179 \\
                        & SR & \textbf{0.585} ($+134\%$) & \textbf{0.483} ($+171\%$) \\
\midrule
\multicolumn{2}{l}{$\text{Base}=0$: Recovery rate\footnotemark[1]}        & 46.5\% & 38.3\% \\
\multicolumn{2}{l}{\phantom{$\text{Base}=0$: }New false-alarm rate\footnotemark[2]}  & 0.73\% & 0.45\% \\
\midrule
\multicolumn{2}{l}{$\text{Base}=1$: Hit-retention rate\footnotemark[3]}       & 94.5\% & 94.7\% \\
\multicolumn{2}{l}{\phantom{$\text{Base}=1$: }Hit-loss rate\footnotemark[4]} & 5.5\% & 5.3\% \\
\multicolumn{2}{l}{\phantom{$\text{Base}=1$: }False-alarm removal rate\footnotemark[5]} & 24.2\% & 27.9\% \\
\bottomrule
\end{tabular*}
\footnotetext[1]{Recovery rate $= P(\text{SR}{=}1\mid \text{Base}{=}0,\text{GT}{=}1)$: heavy-rain pixels missed by exPreCast that SR recovers.}
\footnotetext[2]{New false-alarm rate $= P(\text{SR}{=}1\mid \text{Base}{=}0,\text{GT}{=}0)$: pixels below the threshold that exPreCast correctly leaves unflagged and SR alarms on. Conditioned on $\text{Base}{=}0$, this is not the conventional false-alarm ratio.}
\footnotetext[3]{Hit-retention rate $= P(\text{SR}{=}1\mid \text{Base}{=}1,\text{GT}{=}1)$: true detections preserved by SR.}
\footnotetext[4]{Hit-loss rate $= P(\text{SR}{=}0\mid \text{Base}{=}1,\text{GT}{=}1) = 100\% - \text{hit-retention rate}$: true detections deleted by SR.}
\footnotetext[5]{False-alarm removal rate $= P(\text{SR}{=}0\mid \text{Base}{=}1,\text{GT}{=}0)$: false detections deleted by SR.}
\end{table}

\subsection{Application to another radar dataset: MeteoNet}\label{sec:results-meteonet}

We further evaluate the proposed framework on MeteoNet~\cite{meteonet}, which provides radar observations from a different radar network and climatological regime in France. Unlike the KMA data, which are represented as rainfall intensity, MeteoNet reports radar reflectivity in dBZ. The MeteoNet observations and deterministic backbone forecasts both have a spatial resolution of 1\,km. Consequently, the diffusion module operates entirely at the native resolution and learns a probabilistic residual correction without performing spatial downscaling. This experiment therefore tests whether the benefits of residual diffusion persist when no super-resolution step is involved.

\begin{figure}[!htbp]
\centering
\includegraphics[width=\textwidth]{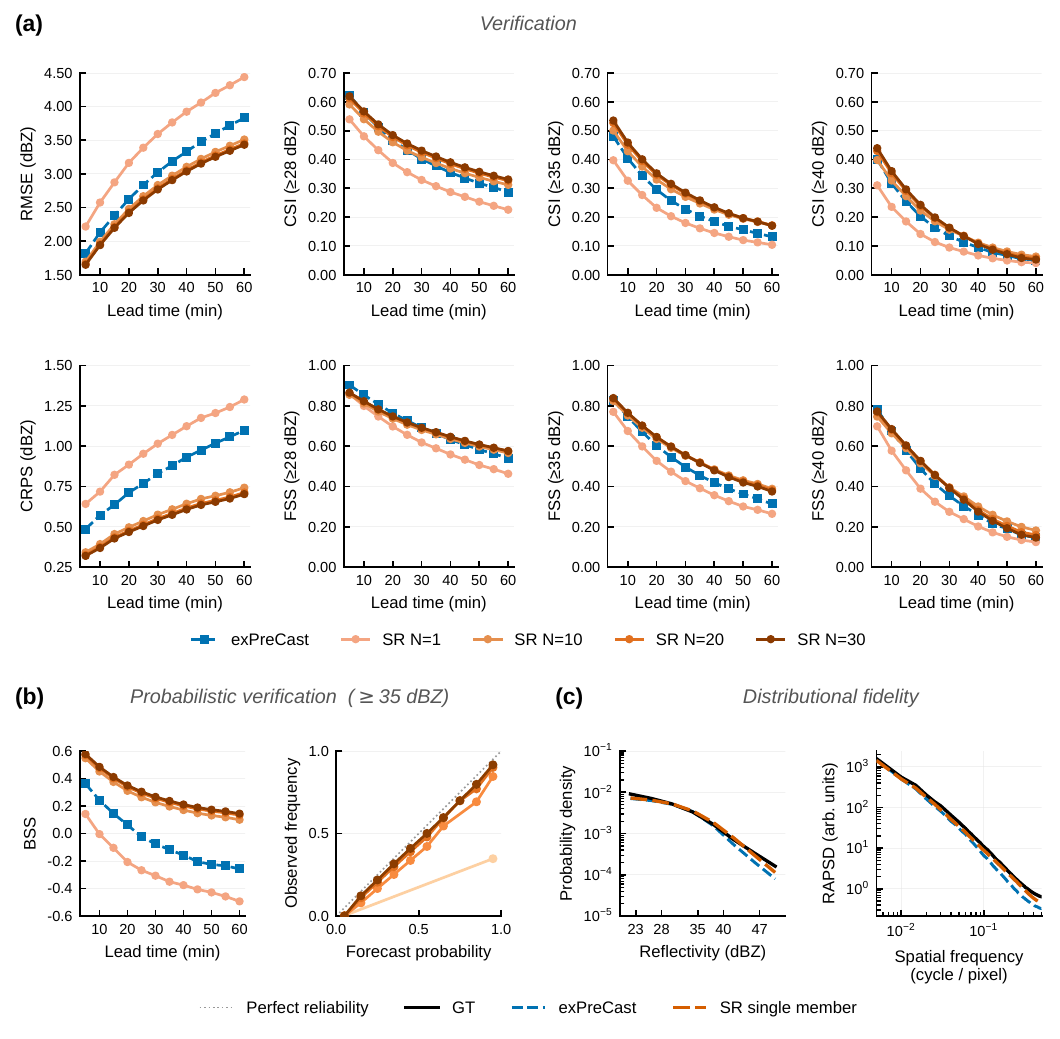}
\caption{\textbf{Quantitative evaluation on the MeteoNet dataset.} Panels and
conventions follow Fig.~\ref{fig:kma-eval}, with two differences: the backbone and the diffusion module both operate at 1\,km, so no pooling is applied and no super-resolution is involved; and thresholds are in dBZ. Scores are computed over $2{,}000$ sequences sampled from the test set in proportion to their peak-intensity distribution (Section~\ref{sec:methods-eval}). (a)~Skill on the 1\,km grid as a function of lead time, with one curve per ensemble size $N\in\{1,10,20,30\}$: RMSE and CRPS; CSI and FSS at the $28$, $35$, and $40$\,dBZ thresholds. FSS uses a $3\times3$ neighborhood ($3\,\text{km}$ at this resolution). The $P\geq 0.3$ alarm criterion is the value selected on KMA (Supplementary Fig. 1), applied here without
adjustment. (b)~Probabilistic skill at the $35$\,dBZ threshold: BSS as a function of lead time, with one curve per ensemble size, and the reliability diagram.
(c)~Reflectivity PDF and RAPSD for the ground truth (solid black), exPreCast (dashed blue), and one individual diffusion member (dashed orange), shown as medians over the test period.}\label{fig:meteonet}
\end{figure}

Because the deterministic MeteoNet backbone already operates at the target resolution, its reflectivity distribution and spatial power spectrum are closer to those of the radar observations than in the KMA 4\,km-to-1\,km setting. Accordingly, the individual diffusion forecasts yield only modest additional agreement in the probability density function and radially averaged power spectral density, as shown in Figure~\ref{fig:meteonet}(c). Nevertheless, the diffusion module remains useful for representing forecast uncertainty and correcting errors in the deterministic prediction.

All model components are trained independently on MeteoNet. Specifically, a MeteoNet-specific deterministic backbone is first trained and then frozen, after which the residual diffusion module is trained using the MeteoNet forecasts and observations. No model weights are transferred from the KMA experiments. The MeteoNet results should therefore be interpreted as an independent replication of the framework rather than as zero-shot cross-dataset transfer. The architecture and training procedure otherwise follow those used for KMA, except that a self-attention block is retained at the U-Net bottleneck because the smaller MeteoNet domain makes its computational cost manageable; further details are provided in the Methods.

Verification follows the KMA protocol, except that no spatial pooling is required because both the backbone and diffusion forecasts are evaluated directly on the 1\,km grid. Scores are computed over $2{,}000$ sequences sampled from the test set in proportion to their peak-intensity distribution (Section~\ref{sec:methods-eval}). Dataset-specific reflectivity thresholds in dBZ are used for the categorical and probabilistic metrics, as described in Methods. As the ensemble size increases, deterministic and probabilistic skill improve consistently across lead times. The reliability curves also move closer to the diagonal, indicating improved calibration of the ensemble exceedance probabilities; the complete numerical results are reported in Supplementary Table 5 and summarized in Figure~\ref{fig:meteonet}(a,b).

The ensemble-mean also remains systematically distinct from the deterministic MeteoNet backbone. Even after the ensemble-size results have stabilized, coherent positive and negative residual structures remain between the 30-member mean and the backbone forecast, as shown in Supplementary Fig. 2. This result indicates that the learned residual distribution has a non-zero conditional mean even when the input and output resolutions are identical. Taken together, the MeteoNet experiments show that the improvements in probabilistic skill and the structured correction of deterministic forecasts are not specific to spatial super-resolution. Rather, residual diffusion provides a general mechanism for converting a deterministic nowcasting model into a probabilistic forecaster at either a finer or unchanged spatial resolution.

\section{Discussion}\label{sec:discussion}

By augmenting exPreCast with a residual diffusion module, we convert a deterministic nowcaster into a probabilistic forecasting framework that generates precipitation ensembles at 1\,km resolution. Our results reveal two complementary roles of the learned residual distribution. Its conditional mean provides a structured correction to the deterministic forecast, whereas its member-to-member variability represents plausible fine-scale precipitation evolution. The framework achieves these improvements with modest additional computational cost. Here, we discuss the interpretation of these two roles and their implications for operational nowcasting.


A residual diffusion module may be viewed simply as adding stochastic fine-scale details to a deterministic prediction. Under this interpretation, averaging sufficiently many samples would remove the stochastic component and recover the deterministic backbone. Our results show that this interpretation is incomplete. Even after the ensemble statistics have largely stabilized, the 30-member mean remains distinct from the exPreCast forecast and achieves better verification scores. The learned residual distribution therefore contains a predictable, non-zero-mean component in addition to stochastic variability.

This behavior arises because the frozen deterministic backbone is not assumed to represent the conditional mean under all the information available to the diffusion module. In addition to the exPreCast forecast sequence, the diffusion module receives preceding radar observations that describe the recent evolution of the precipitation system. This temporal context provides information from which lead-time-dependent displacement and intensity errors in the backbone may be inferred. The degradation observed when past radar observations are removed supports this interpretation (Supplementary Fig. 3).

The ensemble-mean correction is also spatially structured. The small net change in domain-integrated precipitation, combined with the much larger absolute difference between the backbone and the ensemble-mean, indicates that positive and negative corrections largely offset one another over the domain. The diffusion module therefore does not merely increase or decrease the total predicted rainfall. Instead, it redistributes precipitation by modifying the location, extent, intensity, and organization of the predicted rain system. This distinction is important because many consequential nowcasting errors arise from spatial displacement or excessive smoothing rather than from a uniform domain-wide intensity bias.

Residual diffusion has previously been applied to static atmospheric downscaling~\cite{corrdiff}. In the present forecasting setting, it instead acts as a temporal probabilistic post-processor that corrects predictable components of forecast error while representing unresolved future variability. Individual members and the ensemble-mean consequently serve different purposes. An individual member preserves localized extremes and fine-scale structures but is affected by stochastic positional variability. The ensemble-mean provides a more stable estimate of expected precipitation but smooths extremes that occur at different locations across members. Exceedance probabilities and alarm masks retain information about the occurrence of hazardous rainfall without averaging its intensity directly. No single one of these products is universally preferable; their usefulness depends on whether the application requires a representative realization, an expected precipitation field, or a probability of threshold exceedance.

The probabilistic extension is compact relative to the deterministic backbone. In the KMA configuration, the diffusion module contains $6.00$\,M parameters, about one-fifth of the $32.0$\,M exPreCast backbone~\cite{exPreCast}. Because exPreCast remains frozen, converting it into a probabilistic forecaster requires training only the additional residual module. The results demonstrate that this relatively small module is sufficient to produce meaningful probabilistic variability and structured ensemble-mean corrections.

On a single NVIDIA A6000 GPU, the diffusion module generates one member of a 1\,h forecast in $3.4$ seconds and a 30-member ensemble sequentially in approximately $100$ seconds. Ensemble members are generated independently given the conditioning inputs, so sampling can in principle be parallelized across multiple GPUs or accelerated through batched inference. These runtimes make the framework compatible with frequently updated nowcasting, although a complete operational assessment would also need to account for radar ingestion, preprocessing, backbone inference, and dissemination of forecast products.

The probabilistic output provides flexibility in translating the forecast into operational decisions. For a rainfall-intensity threshold $\tau$, the exceedance probability is estimated as the fraction of ensemble members predicting rainfall of at least $\tau$. In the main experiments, an alarm is issued when this probability is at least $0.3$. This probability criterion should not be interpreted as universally optimal. The skill-maximizing value varies with rainfall intensity and forecasting region, as shown in Supplementary Fig. 1. In practice, the criterion could be selected using regional validation data or a cost--loss analysis reflecting the relative consequences of missed events and false alarms. The underlying probability field can therefore support different warning strategies without retraining the forecasting model. In this work we fixed the criterion at a single value chosen for heavy-rain intensities. The same value is used for every threshold and both datasets, so all results reported here rest on one setting. In an operational setting the criterion could instead be adjusted to the intensity threshold and the region at hand. This would give better skill than the fixed value used here.

The probability field itself remains imperfectly calibrated. Although BSS and reliability improve consistently with ensemble size, the rank histogram in Supplementary Fig. 4 indicates that some under-dispersion remains. Under-dispersion is a known tendency of residual diffusion: CorrDiff reports the same behavior, with an ensemble spread too small relative to the ensemble-mean error~\cite{corrdiff}. Whether it follows from the residual formulation itself, from the frozen backbone, or from the sampler is not resolved by the present experiments. Calibration-aware training or post-processing therefore provides a natural direction for future work, particularly for applications involving rare and high-impact rainfall thresholds.


The current experiments focus on two radar datasets and use exPreCast as the deterministic backbone. The consistent behavior observed on KMA and MeteoNet---including the setting in which no resolution enhancement is required---is encouraging and suggests that the framework is not specific to a single radar network or spatial-resolution setting. Evaluating the approach with additional deterministic nowcasters, radar networks, climatic regimes, and longer forecast horizons would further establish its broader applicability. It would also be useful to investigate whether a residual diffusion module trained in one setting can be efficiently adapted to another, rather than trained separately for each dataset.

Overall, these results establish residual diffusion as an effective and computationally practical extension of deterministic precipitation nowcasting. Further advances in calibration and evaluation across a broader range of forecasting systems could strengthen its potential for operational probabilistic forecasting.

\section{Methods}\label{sec:methods}
\subsection{Problem formulation and framework overview}
\label{sec:methods-problem}
exPreCast-ENS combines a deterministic nowcasting backbone with a residual diffusion model. The backbone first predicts the future radar sequence, after which the diffusion model generates probabilistic corrections conditioned on both the deterministic forecast and the preceding observations. This decomposition preserves the predictive structure of the backbone while allowing multiple plausible realizations of its forecast error.

We write $\mathbf{x}^{\mathrm{in}}_{t-L+1:t}:=(x^{\mathrm{in}}_{t-L+1},\ldots,x^{\mathrm{in}}_{t})$ for the sequence of $L$ past radar fields on the backbone input grid, and $\mathbf{y}:=(y_{t+1},\ldots,y_{t+K})$ for the $K$ future fields on the target grid. On KMA the input grid is 4\,km and the target grid 1\,km; on MeteoNet both are 1\,km. We use exPreCast~\cite{exPreCast} as the deterministic backbone, and write its output as $\hat{y}_{\mathrm{det}}=f_{\mathrm{det}}(\mathbf{x}^{\mathrm{in}}_{t-L+1:t};\theta_{\mathrm{det}})$, where $\theta_{\mathrm{det}}$ denotes backbone parameters. The backbone is a UNet-style neural network built on 3D Swin Transformer and trained with the Fourier Amplitude Correlation Loss (FACL). A separate backbone is trained for each dataset using the splits described in Section~\ref{sec:methods-data}, and its weights are frozen while training the diffusion module.

The residual and the diffusion model are defined on the target grid. Let $\mathcal{I}$ denote bilinear interpolation from the backbone grid to the
target grid; on KMA this maps 4\,km fields to 1\,km, and on MeteoNet, where
the two grids coincide, $\mathcal{I}$ reduces to the identity map. The residual is then defined as
\begin{equation}\label{eq:residual}
    r := \mathbf{y} - \mathcal{I}(\hat{y}_{\mathrm{det}}).
\end{equation}
The diffusion module learns the conditional distribution $p_{\theta}\!\left(r \mid \mathcal{I}(\hat{y}_{\mathrm{det}}),\, \mathcal{I}(\mathbf{x}^{\mathrm{in}}_{t-L+1:t})\right)$. Specifically, the deterministic output $\hat{y}_{\mathrm{det}}$ and the past observation sequence $\mathbf{x}^{\mathrm{in}}_{t-L+1:t}$ are concatenated along the channel dimension to form $K+L$ conditioning channels, which the denoising network takes together with the $K$ noisy residual channels. Providing the past sequence directly retains temporal information that may not be fully represented in the deterministic forecast. Indeed, we observe in Supplementary Fig. 3 that removing the past sequence from the input degrades skill scores.

At inference, each sampled residual $\hat{r}^{(n)}$ is added to the deterministic forecast to form an ensemble member,
\begin{equation}
    \hat{y}^{(n)} := \mathcal{I}(\hat{y}_{\mathrm{det}}) + \hat{r}^{(n)},
    \qquad n=1,\ldots,N.
    \label{eq:ensemble-member}
\end{equation}
The resulting ensemble represents multiple plausible corrections to the deterministic forecast and characterizes the uncertainty captured by the residual diffusion model.

\subsection{Residual diffusion module}\label{sec:methods-diffusion}
We implement the conditional residual model described above using the EDM formulation~\cite{edm}, following its use for residual diffusion in CorrDiff~\cite{corrdiff}. The standardized residual is perturbed with Gaussian noise across a range of levels, and the denoising network is trained to recover it under the EDM preconditioning and loss weighting. Writing $r_0$ for the standardized residual and $c=(\mathcal{I}(\hat{y}_{\mathrm{det}}), \mathcal{I}(\mathbf{x}^{\mathrm{in}}_{t-L+1:t}))$ for the conditioning channels, a noise level $\sigma$ is drawn as $\ln\sigma \sim \mathcal{N}(P_{\mathrm{mean}}, P_{\mathrm{std}}^{2})$ with $P_{\mathrm{mean}}=-1.2$ and $P_{\mathrm{std}}=1.2$, and a noisy residual is formed as $r_\sigma = r_0 + \sigma\epsilon$ with $\epsilon\sim\mathcal{N}(0,I)$.

The denoising network $F_\theta$ is wrapped in the EDM preconditioning,
\begin{equation}
D_\theta(r_\sigma, \sigma, c) = c_{\mathrm{skip}}(\sigma)\,r_\sigma
+ c_{\mathrm{out}}(\sigma)\,F_\theta\!\bigl(c_{\mathrm{in}}(\sigma)\,r_\sigma,\,
c_{\mathrm{noise}}(\sigma),\, c\bigr),
\label{eq:edm-precond}
\end{equation}
with the coefficients of ref.~\cite{edm},
\begin{equation}
c_{\mathrm{skip}}(\sigma)=\frac{\sigma_{\mathrm{data}}^{2}}{\sigma^{2}+\sigma_{\mathrm{data}}^{2}},
\quad
c_{\mathrm{out}}(\sigma)=\frac{\sigma\,\sigma_{\mathrm{data}}}{\sqrt{\sigma^{2}+\sigma_{\mathrm{data}}^{2}}},
\quad
c_{\mathrm{in}}(\sigma)=\frac{1}{\sqrt{\sigma^{2}+\sigma_{\mathrm{data}}^{2}}},
\quad
c_{\mathrm{noise}}(\sigma)=\tfrac{1}{4}\ln\sigma,
\label{eq:edm-coeff}
\end{equation}
and $\sigma_{\mathrm{data}}=1.0$, matching the unit standard deviation of the standardized residual. The network is trained to recover $r_0$ under the EDM-weighted loss
\begin{equation}\label{eq:edm-loss-main}
\mathcal{L}(\theta) = \mathbb{E}_{r_0,\sigma,\epsilon}\!\left[\lambda(\sigma)\,\bigl\lVert D_\theta(r_\sigma,\sigma,c) - r_0\bigr\rVert_2^{2}\right],
\qquad \lambda(\sigma) = \frac{\sigma^{2}+\sigma_{\mathrm{data}}^{2}}{(\sigma\,\sigma_{\mathrm{data}})^{2}},
\end{equation}
whose weighting follows from the same coefficients. This preconditioned parameterization suits the wide dynamic range of precipitation residuals and supports few-step deterministic sampling without distillation. The noise distribution $(P_{\mathrm{mean}}, P_{\mathrm{std}})$ follows the EDM defaults.

At training, the residual $r$ is standardized channel-wise using fixed mean and standard deviation values computed once on the training set. At inference, the sampler generates a standardized residual, which is denormalized and added to $\mathcal{I}(\hat{y}_{\mathrm{det}})$. The diffusion module operates in radar reflectivity for both datasets. After denormalization and reconstruction, KMA forecasts are converted to rainfall rate for evaluation, whereas MeteoNet forecasts are evaluated directly in reflectivity (dBZ).

\paragraph*{Denoising network.}
$F_\theta$ is a U-Net operating on the full domain at 1\,km resolution. Its input concatenates the $K$ noisy residual channels with the $K+L$ conditioning channels along the channel dimension ($6{+}13$ on KMA, $12{+}24$ on MeteoNet); all channels share the 1\,km grid, so on KMA the 4\,km backbone forecast and past sequence are bilinearly interpolated beforehand (Section~\ref{sec:methods-problem}) and the network input and output reside on the same grid. The network has four stages (three encoder stages and a bottleneck), base channel width $64$, and channel widths $\{64, 128, 256, 256\}$ from top to bottom. Each stage is a single residual block of two $3\times3$ convolutions with replicate padding, GroupNorm, and SiLU activations; the noise-level embedding $c_{\mathrm{noise}}(\sigma)$ is injected into each block through a linear projection added after the first convolution.

The bottleneck stage contains a spatial multi-head self-attention block (4 heads), standard practice in diffusion-based generative architectures~\cite{edm,corrdiff}. On the KMA domain ($1024\times1024$ at 1\,km) the bottleneck feature map is large enough that this attention exceeds the memory budget of a single GPU under our training configuration, so we bypass it (acting as identity) during both training and inference on KMA; on the smaller MeteoNet domain it fits and is retained. The network therefore contains $6.00$\,M parameters in the KMA configuration and $6.28$\,M in the MeteoNet configuration. This bypass is an engineering compromise driven by single-GPU memory rather than a design preference: a larger memory budget, or domain patching at the cost of seam-handling, would allow the attention to be retained throughout.

\paragraph*{Sampling.}
Samples are drawn with Heun's second-order deterministic sampler over $M=20$ steps on a Karras-type schedule,
\begin{equation}
\sigma_i = \Bigl(\sigma_{\max}^{1/\rho}
+ \tfrac{i}{M-1}\bigl(\sigma_{\min}^{1/\rho} - \sigma_{\max}^{1/\rho}\bigr)\Bigr)^{\rho},
\qquad i = 0,\dots,M-1,
\label{eq:karras}
\end{equation}
with $\sigma_{\min}=0.002$, $\sigma_{\max}=80$, and $\rho=7$, following ref.~\cite{edm}. The schedule parameters $(\sigma_{\min}, \sigma_{\max}, \rho)$ follow the EDM defaults; the step count $M=20$ is our choice. Each sampler call is initialized with independent Gaussian noise and returns one standardized residual, which is denormalized and added to $\mathcal{I}(\hat{y}_{\mathrm{det}})$ to give a precipitation forecast $\hat{y}^{(n)}=\mathcal{I}(\hat{y}_{\mathrm{det}})+\hat{r}^{(n)}$. We draw $30$ realizations per test case and report results for $N\in\{1,10,20,30\}$ using the first $N$ of them, so the ensembles are nested. Evaluating each of the $30$ members individually gives a spread that is small relative to their margin over the backbone (Supplementary Fig. 5), so the $N=1$ results do not depend on the member drawn.

\paragraph*{Optimization.}
We use the AdamW optimizer with a peak learning rate of $1\times10^{-4}$ and a cosine schedule with $1{,}000$ warmup steps over $400{,}000$ training steps. Training is in mixed precision (FP16) with gradient scaling, and gradient norms are clipped at $1.0$. We keep an exponential moving average (EMA) of the network weights with decay $0.999$ and use the EMA weights at inference.

\paragraph*{Alarm mask.}
In addition to the ensemble-mean, we introduce an \textit{alarm mask} to summarize the potential occurrence of precipitation exceeding a prescribed intensity threshold. For an intensity threshold $\tau$, the ensemble exceedance probability is computed separately at each grid point and lead time as
\begin{equation}
    \hat{P}(y \ge \tau):= \frac{1}{N}\sum_{n=1}^{N}\mathbf{1}\left[\hat{y}^{(n)} \ge \tau\right],
\end{equation}
where $\mathbf{1}[\cdot]$ denotes the indicator function. We then define the proposed alarm mask as
\begin{equation}
    M_{\mathrm{alarm}}(\tau) := \mathbf{1}\left[\hat{P}(y \ge \tau) \ge p_{\mathrm{alarm}}\right],
\end{equation}
where $p_{\mathrm{alarm}}$ is the probability criterion for issuing an alarm. Thus, $M_{\mathrm{alarm}}(\tau)$ is a binary field indicating locations where the ensemble assigns sufficient probability to an exceedance of $\tau$. We set $p_{\mathrm{alarm}}=0.3$ in all experiments; the selection of this value is examined in Supplementary Fig. 1.

\subsection{Datasets and splits}\label{sec:methods-data}
\paragraph*{KMA radar.}
The KMA radar dataset consists of radar reflectivity composites covering the Korean Peninsula, provided at 4\,km resolution for the deterministic backbone input and at 1\,km resolution for the target. The temporal cadence is 10 minutes, resulting in 7-frame input and 6-frame forecast sequences. The 1\,km target grid covers a $1024\times1024$ domain centered on the Korean Peninsula. We adopt the train/validation/test split of exPreCast~\cite{exPreCast}: 2014--2021 for training, 2022 for validation, and 2023 for testing.

\paragraph*{MeteoNet.}
We use the MeteoNet radar dataset~\cite{meteonet}, which provides radar reflectivity composites at 1\,km resolution over a $416\times416$ regional domain in southeastern France. We retain the native 5-minute temporal cadence, resulting in 12-frame input and forecast sequences. We use the same train/validation/test split as the MeteoNet evaluation in exPreCast~\cite{exPreCast}. Because the input and target both reside at 1\,km, the diffusion module performs probabilistic residual correction without a change in spatial resolution.

\subsection{Evaluation metrics}\label{sec:methods-eval}

\paragraph*{Evaluation sample and domain.}
For KMA, all scores are computed over the full 2023 test period (52{,}325 forecast initializations). For MeteoNet, we evaluate $2{,}000$ of the $59{,}408$ available sequences. Sequences are grouped by their peak reflectivity and drawn from each group in proportion to its size, so the sample preserves the intensity distribution of the full test set rather than enriching intense cases. KMA scores are accumulated over the radar-coverage mask (41{,}572 of 65{,}536 cells on the 4\,km grid); MeteoNet has no such mask and is evaluated over the full domain. All results are reported as a function of lead time (KMA: T+10--T+60\,min, 6 frames; MeteoNet: T+5--T+60\,min, 12 frames).

\paragraph*{Deterministic skill.}
We report the root-mean-square error (RMSE), the Critical Success Index (CSI), the Probability of Detection (POD), and the Fractions Skill Score (FSS) at intensity thresholds of $10$, $20$, and $40$ ($\mathrm{mm}\,\mathrm{h}^{-1}$) for KMA and $28$, $35$, and $40$ (dBZ) for MeteoNet. For the ensemble, the threshold-based scores are computed from the alarm mask $M_{\mathrm{alarm}}(\tau)$ of Section~\ref{sec:methods-diffusion}, that is, at the $p_{\mathrm{alarm}}=0.3$ exceedance criterion. CSI and POD are computed pixel-wise; FSS uses a $3\times3$ neighborhood, corresponding to $12$\,km on the 4\,km verification grid.

\paragraph*{Probabilistic skill.}
We report the Continuous Ranked Probability Score (CRPS), the Brier Skill Score (BSS), and reliability diagrams. BSS and the reliability diagrams use the ensemble exceedance probability $\hat{P}(y \ge \tau)$ of Section~\ref{sec:methods-diffusion} as the forecast probability, rather than the binarized alarm mask. BSS is referenced to the sample climatology, $\mathrm{BS}_{\mathrm{clim}}=\bar{o}(1-\bar{o})$, where $\bar{o}$ is the observed event frequency over the evaluated pixels. For the deterministic baseline, CRPS reduces by definition to the mean absolute error (MAE), and we report this MAE in the CRPS column for direct comparison with ensemble outputs.

\paragraph*{Grid alignment.}
Comparisons between the 1\,km SR ensemble and the 4\,km exPreCast baseline (Fig.~\ref{fig:kma-eval}) are performed on the 4\,km verification grid. Predictions at 1\,km are upscaled to the 4\,km cell using mean pooling for RMSE and CRPS, and max pooling for threshold-based metrics (CSI, POD, FSS) and for the exceedance probability underlying BSS and the reliability diagrams, so that a 4\,km cell counts as an event whenever the threshold is exceeded anywhere within it. Analyses of the rainfall distribution and spatial spectral characteristics are performed on the 1\,km grid, with bilinear interpolation of exPreCast as a reference for the resolution change.

\paragraph*{Distributional metrics.}
The rainfall probability density function (PDF) and the radially averaged power spectral density (RAPSD) are evaluated from a single ensemble member rather than the ensemble-mean, since averaging members suppresses fine-scale variability and the mean field would not represent the distribution the model generates. RAPSD is the radial mean of $\lvert\mathcal{F}\{x\cdot w\}\rvert^{2}$, where $\mathcal{F}$ is the 2-D discrete Fourier transform and $w$ a 2-D Hann window, taken over annuli of constant wavenumber, so the ordinate is power per Fourier mode. The transform is unnormalized and the window energy is not compensated. All fields are processed identically on a common grid, and only relative differences between curves are interpreted.

\paragraph*{Conditional verification.}
For the event-level analysis in Section~\ref{sec:results-cases} we additionally report five conditional rates that compare the SR alarm mask with the deterministic baseline, conditioned on the baseline outcome. Writing $\mathrm{Base}$ and $\mathrm{SR}$ for the two binary forecast fields
and $\mathrm{GT}$ for the binary observation,
\begin{align*}
\text{recovery rate} &= \Pr(\mathrm{SR}{=}1 \mid \mathrm{Base}{=}0,\ \mathrm{GT}{=}1), \\
\text{new false-alarm rate} &= \Pr(\mathrm{SR}{=}1 \mid \mathrm{Base}{=}0,\ \mathrm{GT}{=}0),\\
\text{hit-retention rate} &= \Pr(\mathrm{SR}{=}1 \mid \mathrm{Base}{=}1,\ \mathrm{GT}{=}1), \\
\text{hit-loss rate} &= \Pr(\mathrm{SR}{=}0 \mid \mathrm{Base}{=}1,\ \mathrm{GT}{=}1),\\
\text{false-alarm removal rate} &= \Pr(\mathrm{SR}{=}0 \mid \mathrm{Base}{=}1,\ \mathrm{GT}{=}0). &&
\end{align*}
The first two isolate what the ensemble adds where the baseline issued no alarm; the last three, what it does with the alarms the baseline already had. Hit-retention and hit-loss share a denominator and sum to one. The qualifier ``new'' in the second is deliberate: it is conditioned on $\mathrm{Base}{=}0$ and is therefore not the conventional false-alarm ratio. 
All five are computed on the 4\,km verification grid at the $10\,\mathrm{mm}\,\mathrm{h}^{-1}$ threshold and the $p_{\mathrm{alarm}}=0.3$ criterion, after upscaling the 1\,km output to the 4\,km grid via max pooling.
Full-period values for 2023 across all lead times and thresholds ($10,20,40$ $\mathrm{mm}\,\mathrm{h}^{-1}$) are reported in Supplementary Table 4.
\subsection{Hardware and inference cost}\label{sec:methods-impl}
Training is performed on $4\times$ NVIDIA RTX A6000 GPUs using PyTorch DistributedDataParallel; the KMA module is trained for approximately 4 days (${\sim}16$ GPU-days) and the MeteoNet module for approximately 7 days (${\sim}28$ GPU-days). 

At inference on a single A6000, the deterministic backbone produces one forecast in 22\,ms on KMA and 74\,ms on MeteoNet. Each stochastic member is generated with 20 sampling steps in 3.4\,s on KMA ($1024\times1024$) and 1.3\,s on MeteoNet ($416\times416$). Generating a full 30-member ensemble sequentially requires ${\sim}100$\,s on KMA and ${\sim}35$\,s on MeteoNet. Because ensemble members are sampled independently, inference can be parallelized across members and GPUs.

\section*{Data availability}
The MeteoNet radar dataset is publicly available from M\'{e}t\'{e}o-France~\cite{meteonet}.
The KMA radar composites used in this study are the same data released with
exPreCast~\cite{exPreCast} and are available at
\url{https://drive.google.com/drive/folders/1wNo9dp1q_88kcS7BXpKZ4BPiWIki_TeN}.
The underlying radar observations are provided by the Korea Meteorological
Administration and are subject to its distribution terms.

\section*{Code availability}
The code that supports the findings of this study is available from the corresponding author upon reasonable request and will be deposited in a public repository upon publication. The official exPreCast weights used as the deterministic backbone are available at \url{https://github.com/tony890048/exPreCast}.

\section*{Acknowledgements}
This work was supported by the ASTRA Project through the National Research Foundation (NRF) funded by the Ministry of Science and ICT (No. RS-2024-00440063).

\section*{Author contributions}
D.P.\ conceived the idea, designed the experiments, performed the analysis, and wrote the manuscript. C.S.\ contributed to the conception of the study, developed the exPreCast backbone, and contributed to writing the manuscript. T.C.\ contributed to the backbone. Y.-G.H.\ and Y.H.\ supervised the project and revised the manuscript. All authors reviewed the manuscript.

\section*{Competing interests}
The authors declare no competing interests.


\clearpage

\clearpage

\setcounter{figure}{0}
\setcounter{table}{0}
\renewcommand{\thefigure}{S\arabic{figure}}
\renewcommand{\thetable}{S\arabic{table}}

\section*{Supplementary Information}

 \begin{figure}[!htbp]
\centering
\includegraphics[width=\textwidth]{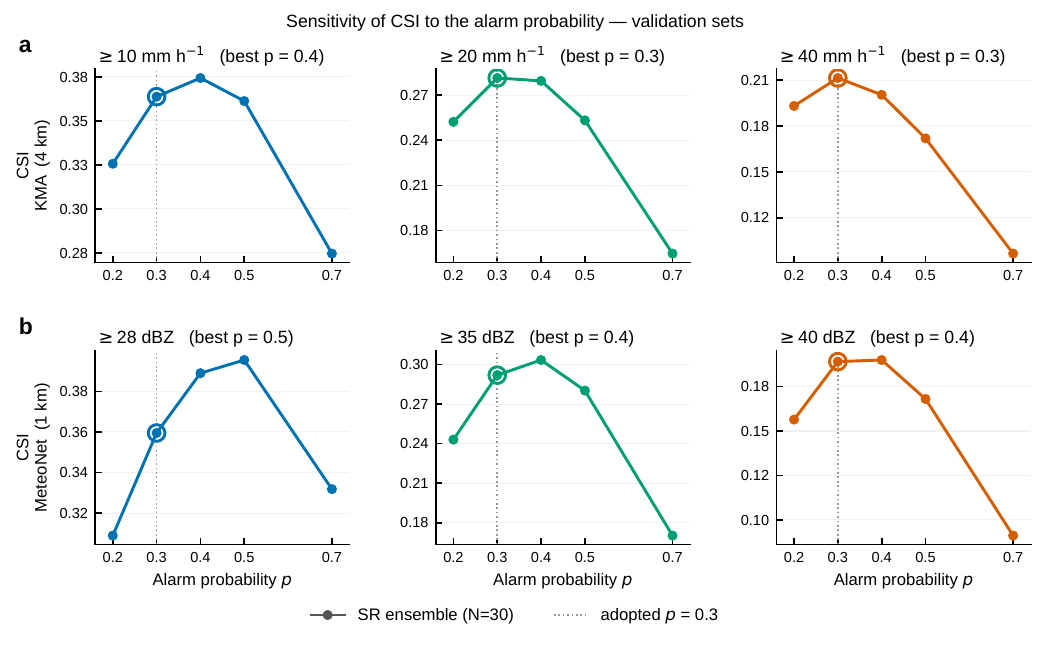}
\caption{\textbf{Sensitivity of CSI to the alarm probability criterion.} Critical Success Index of the 30-member SR ensemble as a function of the alarm probability $p_{\mathrm{alarm}}\in\{0.2,\,0.3,\,0.4,\,0.5,\,0.7\}$, evaluated on the validation split of each dataset (KMA 2022; MeteoNet). (a)~KMA on the 4\,km verification grid at the $10$, $20$, and $40\,\text{mm\,h}^{-1}$ thresholds; (b)~MeteoNet on the 1\,km grid at the $28$, $35$, and $40$\,dBZ thresholds. The value maximizing CSI is given in each panel title, and the value adopted throughout this work, $p_{\mathrm{alarm}}=0.3$, is marked by a vertical dotted line and an open circle. On both datasets the optimum shifts toward lower $p_{\mathrm{alarm}}$ as the intensity threshold increases, since heavy rainfall is exceeded by fewer members and a lower criterion is needed to retain those signals. The optimum on MeteoNet lies above the corresponding KMA value at every threshold, indicating that the skill-maximizing criterion depends on the radar network and climatological regime as well as on intensity. We adopt $p_{\mathrm{alarm}}=0.3$ throughout. On KMA it maximizes CSI at the two higher thresholds, which are the intensities of operational concern. On MeteoNet at the highest threshold, $p_{\mathrm{alarm}}=0.3$ and $0.4$ give almost the same CSI, so we keep $0.3$ for consistency with KMA. At the lower thresholds a different criterion would clearly perform better.}\label{supfig:p-sweep}
\end{figure}

\clearpage

\begin{figure}[!htbp]
\centering
\includegraphics[width=\textwidth]{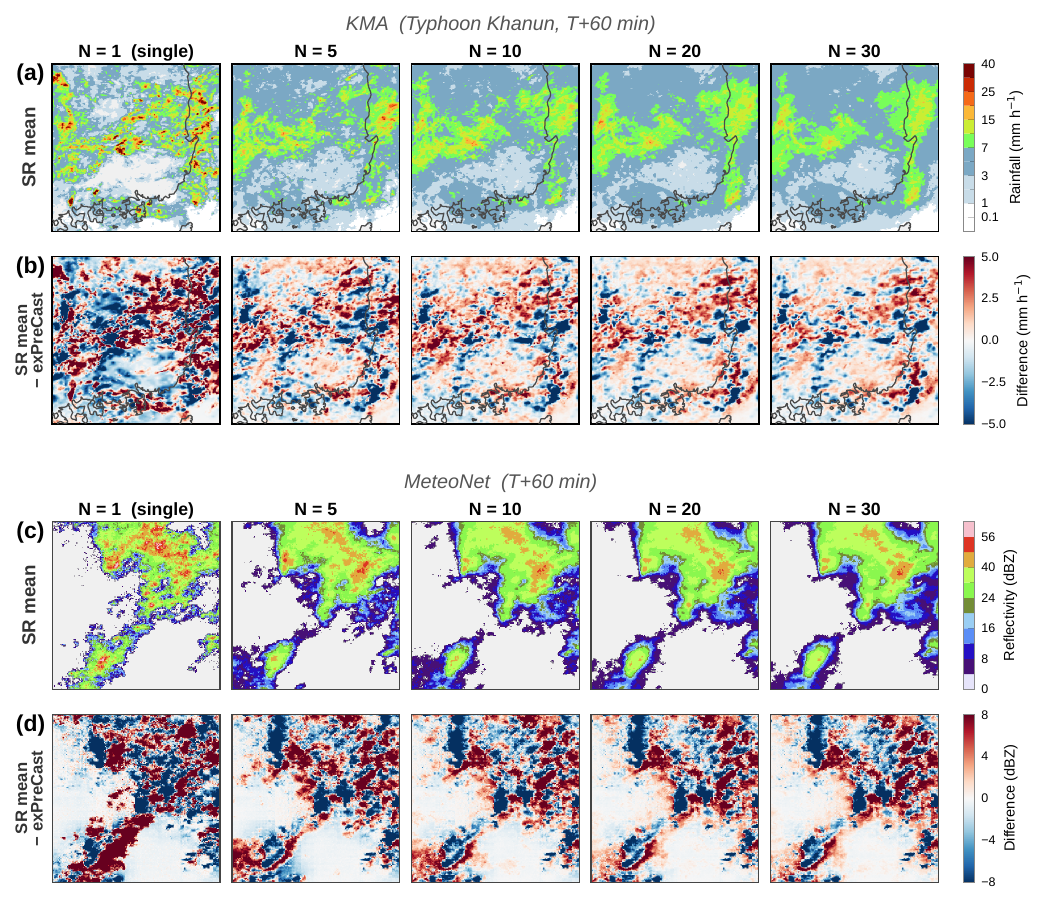}
\caption{\textbf{Persistence of the residual field with ensemble size.} (a,~c)~SR ensemble-mean field and (b,~d)~the residual, the difference between the SR ensemble-mean and exPreCast, for ensemble sizes $N\in\{1,5,10,20,30\}$; the $N=1$ column is a single member. For both KMA (Typhoon Khanun, $T+60$\,min) and MeteoNet ($T+60$\,min), the ensemble-mean residual decreases as the ensemble grows but levels off at a non-zero value, leaving a spatially structured correction field.}\label{supfig:residual}
\end{figure}

\clearpage

\begin{figure}[!htbp]
\centering
\includegraphics[width=\textwidth]{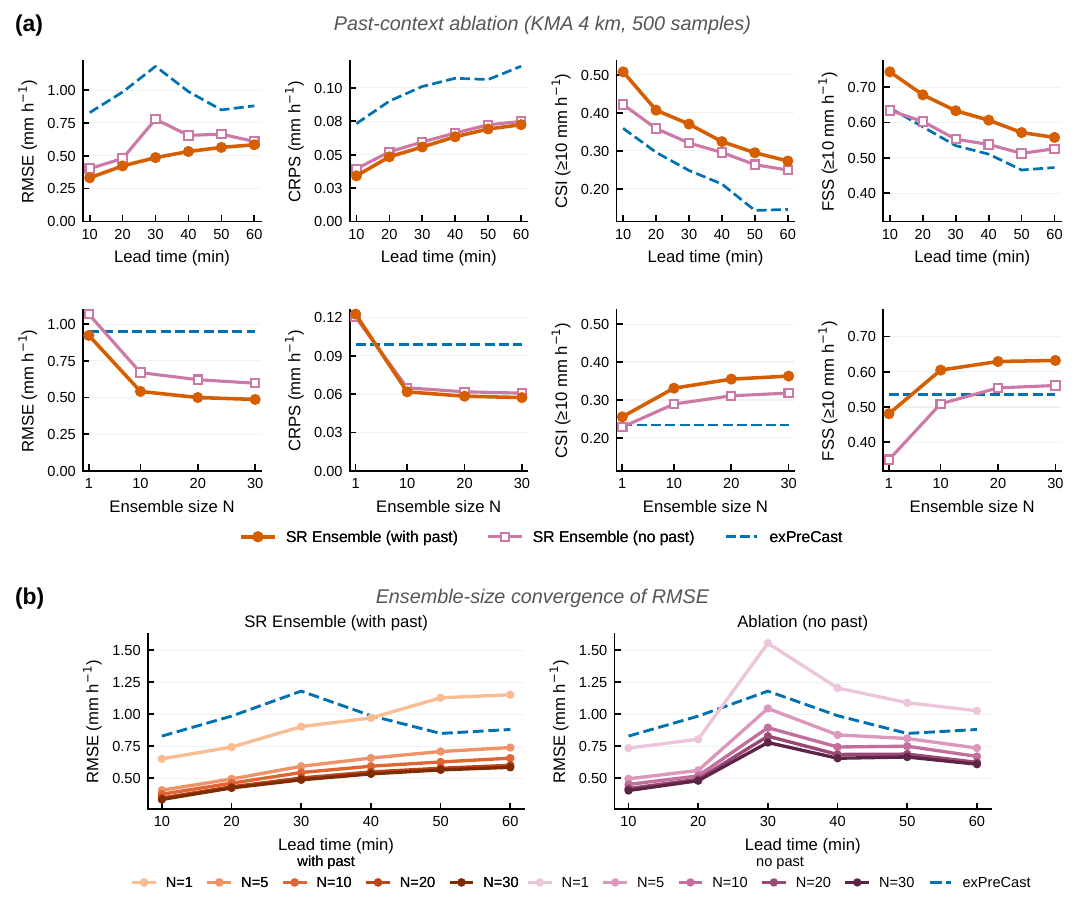}
\caption{\textbf{Past-context ablation (KMA, 4\,km, 500-sample subset stratified by rain fraction).} Effect of removing the past radar observation sequence ($t{-}60$ to $t$\,min) from the diffusion module's conditioning, with the backbone and all other training fixed; the ablated model retains only the backbone forecast, itself a sequence over lead times, as conditioning. (a)~Deterministic and probabilistic metrics (RMSE, CRPS, CSI, FSS at $\geq 10\,\text{mm\,h}^{-1}$) for the SR Ensemble with past context, the ablated model without it, and the exPreCast baseline. Top row: as a function of lead time at fixed ensemble size $N=30$; bottom row: as a function of ensemble size $N$, averaged over lead times. (b)~RMSE as a function of lead time for each ensemble size $N\in\{1,5,10,20,30\}$, with past context (left) and without (right); the exPreCast baseline is shown for reference. Both variants improve over the baseline: the forecast sequence alone already supports a structured correction. At $N=30$ the full model improves on the ablated model across RMSE, CRPS, CSI, and FSS. The gap is clearest where exPreCast's error peaks: in panel~(b) the past-conditioned model suppresses this peak while the ablated model tracks it. Supplying the past observations therefore helps correct the backbone's forecast error.}\label{supfig:past-ablation}
\end{figure}

\clearpage

\begin{figure}[!htbp]
\centering
\includegraphics[width=\textwidth]{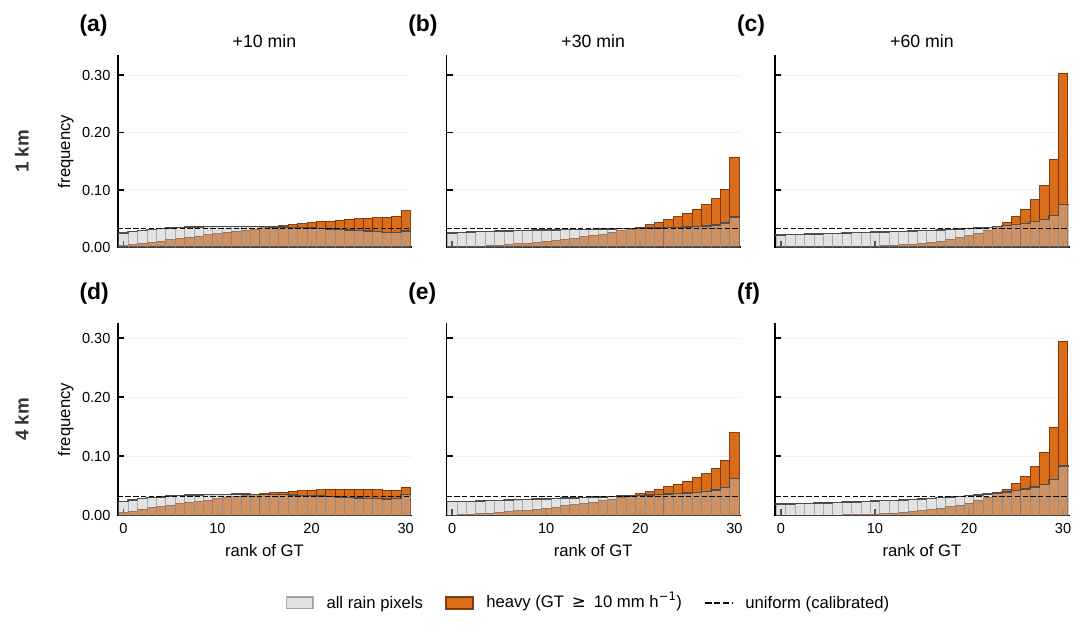}
\caption{\textbf{Rank histograms of the 30-member ensemble, full 2023 test period.} For every (pixel, lead time, sample) the 30 members are sorted and the rank of the ground truth among them is recorded (31 bins, rank $0$--$30$; ties broken by a random uniform offset). Statistics are accumulated over the full KMA 2023 test set (52,325 samples) on the rain mask $(\mathrm{GT}\geq 0.1\,\mathrm{mm\,h^{-1}} \lor \mathrm{exPreCast}\geq 0.1\,\mathrm{mm\,h^{-1}})$ within the radar-coverage domain. Rows: native 1\,km grid (top) and 4\,km grid, obtained by member-wise $4\times4$ average pooling (bottom); the 4\,km rain mask is defined with max-pooled ground truth. Columns: $+10$, $+30$, and $+60$ minute lead times. Gray bars are all rain-mask pixels and orange bars the heavy-rain subset ($\mathrm{GT}\geq 10\,\mathrm{mm\,h^{-1}}$), each normalized to its own total. The dashed line marks the uniform frequency $1/31$ expected from a calibrated ensemble; excess mass in the last bin means the observation exceeded all 30 members. Over all rain-mask pixels the histogram is close to uniform at $+10$\,min and develops a right-hand excess with lead time, reaching $2.3$ times the uniform rate at $+60$\,min. The heavy-rain subset shows the same deficiency far more strongly, reaching $30.4\%$ (1\,km) and $29.5\%$ (4\,km) in the last bin at $+60$\,min while the first bin is essentially empty; conditioning on $\mathrm{GT}\geq 10$ selects large observed values and therefore favors high ranks by construction, so this subset indicates where the deficiency is concentrated rather than establishing it on its own. The 1\,km and 4\,km histograms are nearly identical, so the deficiency reflects the ensemble spread rather than the grid resolution. This is the empirical evidence underlying the under-dispersion discussed in the Discussion.}\label{supfig:rank-histogram}
\end{figure}

\clearpage


\scriptsize
\setlength{\tabcolsep}{4pt}
\renewcommand{\arraystretch}{0.9}
\begin{longtable}{@{}llcccccc@{}}
\caption{\textbf{KMA test set (2023), 4\,km verification grid.}
Deterministic and threshold-based verification at $+10$--$60$\,min lead time. SR Ensemble outputs (1\,km, $N$ members at the $P\geq 0.3$ alarm criterion) are upscaled to the 4\,km baseline grid before comparison (mean pooling for RMSE/CRPS, max pooling for CSI/FSS). FSS is computed with a $3\times 3$ neighborhood window; CSI is computed pixel-wise. Thresholds are in mm\,h$^{-1}$. The ensemble CRPS at $N=1$ is slightly higher than the deterministic baseline (an intrinsic property of single-sample stochastic forecasts) and improves below the baseline once $N\geq 10$. $^{\ast}$For the deterministic baseline (exPreCast), CRPS reduces by definition to MAE; the same value is reported in the CRPS row for direct comparability with the SR Ensemble CRPS. The values in this table correspond to the curves in main-text Fig. 2a of the main text.
}\label{suptab:kma-4km} \\
\toprule
Model & Metric & \multicolumn{6}{c}{Lead time (min)} \\
\cmidrule(lr){3-8}
 & & 10 & 20 & 30 & 40 & 50 & 60 \\
\midrule
\endfirsthead

\multicolumn{8}{l}{\footnotesize\textit{Supplementary Table~\ref{suptab:kma-4km} (continued)}}\\
\toprule
Model & Metric & \multicolumn{6}{c}{Lead time (min)} \\
\cmidrule(lr){3-8}
 & & 10 & 20 & 30 & 40 & 50 & 60 \\
\midrule
\endhead

\midrule \multicolumn{8}{r}{\footnotesize\textit{Continued on next page}}\\
\endfoot

\botrule
\endlastfoot
 \multirow{8}{*}{exPreCast} & RMSE & 1.921 & 2.285 & 2.286 & 2.063 & 2.080 & 2.437 \\*
  & CRPS$^{\ast}$ & 0.0729 & 0.0898 & 0.1003 & 0.1073 & 0.1057 & 0.1159 \\*
  & CSI~$\geq$10 & 0.381 & 0.314 & 0.264 & 0.228 & 0.159 & 0.162 \\*
  & CSI~$\geq$20 & 0.275 & 0.208 & 0.157 & 0.125 & 0.078 & 0.079 \\*
  & CSI~$\geq$40 & 0.185 & 0.126 & 0.086 & 0.064 & 0.039 & 0.038 \\*
  & FSS~$\geq$10 & 0.774 & 0.706 & 0.635 & 0.575 & 0.434 & 0.445 \\*
  & FSS~$\geq$20 & 0.667 & 0.572 & 0.472 & 0.399 & 0.262 & 0.269 \\*
  & FSS~$\geq$40 & 0.546 & 0.432 & 0.321 & 0.253 & 0.156 & 0.160 \\
\midrule
 \multirow{8}{*}{SR $N{=}1$} & RMSE & 1.456 & 1.244 & 1.659 & 1.449 & 1.426 & 1.637 \\*
  & CRPS & 0.0809 & 0.0994 & 0.1185 & 0.1253 & 0.1289 & 0.1382 \\*
  & CSI~$\geq$10 & 0.418 & 0.327 & 0.272 & 0.230 & 0.198 & 0.169 \\*
  & CSI~$\geq$20 & 0.331 & 0.240 & 0.187 & 0.148 & 0.120 & 0.098 \\*
  & CSI~$\geq$40 & 0.227 & 0.153 & 0.112 & 0.081 & 0.063 & 0.047 \\*
  & FSS~$\geq$10 & 0.815 & 0.727 & 0.647 & 0.581 & 0.523 & 0.468 \\*
  & FSS~$\geq$20 & 0.747 & 0.636 & 0.535 & 0.456 & 0.391 & 0.333 \\*
  & FSS~$\geq$40 & 0.622 & 0.501 & 0.396 & 0.315 & 0.253 & 0.201 \\
\midrule
 \multirow{8}{*}{SR $N{=}10$} & RMSE & 0.721 & 0.756 & 0.922 & 0.914 & 0.936 & 0.983 \\*
  & CRPS & 0.0391 & 0.0510 & 0.0607 & 0.0666 & 0.0713 & 0.0759 \\*
  & CSI~$\geq$10 & 0.456 & 0.399 & 0.341 & 0.309 & 0.286 & 0.258 \\*
  & CSI~$\geq$20 & 0.385 & 0.327 & 0.262 & 0.227 & 0.196 & 0.168 \\*
  & CSI~$\geq$40 & 0.283 & 0.240 & 0.181 & 0.146 & 0.113 & 0.088 \\*
  & FSS~$\geq$10 & 0.787 & 0.741 & 0.672 & 0.636 & 0.617 & 0.580 \\*
  & FSS~$\geq$20 & 0.735 & 0.690 & 0.596 & 0.556 & 0.515 & 0.467 \\*
  & FSS~$\geq$40 & 0.629 & 0.604 & 0.498 & 0.452 & 0.384 & 0.321 \\
\midrule
 \multirow{8}{*}{SR $N{=}20$} & RMSE & 0.673 & 0.719 & 0.859 & 0.869 & 0.898 & 0.939 \\*
  & CRPS & 0.0368 & 0.0483 & 0.0574 & 0.0634 & 0.0681 & 0.0725 \\*
  & CSI~$\geq$10 & 0.482 & 0.424 & 0.364 & 0.330 & 0.303 & 0.275 \\*
  & CSI~$\geq$20 & 0.415 & 0.353 & 0.285 & 0.245 & 0.206 & 0.175 \\*
  & CSI~$\geq$40 & 0.313 & 0.265 & 0.202 & 0.156 & 0.113 & 0.083 \\*
  & FSS~$\geq$10 & 0.804 & 0.759 & 0.690 & 0.654 & 0.630 & 0.593 \\*
  & FSS~$\geq$20 & 0.758 & 0.713 & 0.620 & 0.575 & 0.525 & 0.472 \\*
  & FSS~$\geq$40 & 0.662 & 0.635 & 0.529 & 0.466 & 0.378 & 0.301 \\
\midrule
 \multirow{8}{*}{SR $N{=}30$} & RMSE & 0.653 & 0.706 & 0.837 & 0.853 & 0.885 & 0.922 \\*
  & CRPS & 0.0360 & 0.0474 & 0.0564 & 0.0623 & 0.0671 & 0.0713 \\*
  & CSI~$\geq$10 & 0.491 & 0.433 & 0.372 & 0.337 & 0.308 & 0.280 \\*
  & CSI~$\geq$20 & 0.425 & 0.362 & 0.292 & 0.251 & 0.209 & 0.177 \\*
  & CSI~$\geq$40 & 0.323 & 0.273 & 0.208 & 0.158 & 0.111 & 0.080 \\*
  & FSS~$\geq$10 & 0.810 & 0.764 & 0.695 & 0.658 & 0.633 & 0.595 \\*
  & FSS~$\geq$20 & 0.766 & 0.720 & 0.626 & 0.580 & 0.525 & 0.470 \\*
  & FSS~$\geq$40 & 0.673 & 0.643 & 0.537 & 0.467 & 0.369 & 0.287 \\
\end{longtable}

\normalsize
\setlength{\tabcolsep}{6pt}
\renewcommand{\arraystretch}{1.0}
\clearpage


\scriptsize
\setlength{\tabcolsep}{4pt}
\renewcommand{\arraystretch}{0.9}
\begin{longtable}{@{}llcccccc@{}}
\caption{\textbf{KMA test set (2023), 1\,km native verification grid.}
Same metrics as Supplementary Table~\ref{suptab:kma-4km}, but evaluated on the 1\,km grid. exPreCast values are obtained by bilinear interpolation of the 4\,km deterministic forecast to the 1\,km grid for reference only---they are \emph{not} a fair equal-resolution baseline, as bilinear interpolation adds no high-resolution information and inflates RMSE relative to the native 4\,km evaluation. SR Ensemble outputs are at native 1\,km resolution. FSS is computed with a $3\times 3$ neighborhood window; CSI is computed pixel-wise. $^{\ast}$For the deterministic baseline (exPreCast), CRPS reduces by definition to MAE; the same value is reported in the CRPS row for direct comparability with the SR Ensemble CRPS.
}\label{suptab:kma-1km} \\
\toprule
Model & Metric & \multicolumn{6}{c}{Lead time (min)} \\
\cmidrule(lr){3-8}
 & & 10 & 20 & 30 & 40 & 50 & 60 \\
\midrule
\endfirsthead

\multicolumn{8}{l}{\footnotesize\textit{Supplementary Table~\ref{suptab:kma-1km} (continued)}}\\
\toprule
Model & Metric & \multicolumn{6}{c}{Lead time (min)} \\
\cmidrule(lr){3-8}
 & & 10 & 20 & 30 & 40 & 50 & 60 \\
\midrule
\endhead

\midrule \multicolumn{8}{r}{\footnotesize\textit{Continued on next page}}\\
\endfoot

\botrule
\endlastfoot
 \multirow{8}{*}{exPreCast} & RMSE & 1.409 & 1.702 & 1.743 & 1.654 & 1.657 & 1.903 \\*
  & CRPS$^{\ast}$ & 0.0728 & 0.0897 & 0.1003 & 0.1074 & 0.1061 & 0.1161 \\*
  & CSI~$\geq$10 & 0.433 & 0.345 & 0.289 & 0.248 & 0.187 & 0.180 \\*
  & CSI~$\geq$20 & 0.321 & 0.234 & 0.175 & 0.139 & 0.091 & 0.087 \\*
  & CSI~$\geq$40 & 0.204 & 0.132 & 0.087 & 0.065 & 0.041 & 0.039 \\*
  & FSS~$\geq$10 & 0.713 & 0.608 & 0.533 & 0.474 & 0.384 & 0.368 \\*
  & FSS~$\geq$20 & 0.610 & 0.479 & 0.381 & 0.313 & 0.218 & 0.208 \\*
  & FSS~$\geq$40 & 0.458 & 0.318 & 0.221 & 0.170 & 0.112 & 0.105 \\
\midrule
 \multirow{8}{*}{SR $N{=}1$} & RMSE & 2.413 & 1.658 & 2.327 & 1.869 & 1.736 & 2.154 \\*
  & CRPS & 0.0950 & 0.1110 & 0.1287 & 0.1343 & 0.1365 & 0.1454 \\*
  & CSI~$\geq$10 & 0.357 & 0.262 & 0.213 & 0.174 & 0.145 & 0.122 \\*
  & CSI~$\geq$20 & 0.268 & 0.177 & 0.132 & 0.100 & 0.078 & 0.062 \\*
  & CSI~$\geq$40 & 0.170 & 0.101 & 0.069 & 0.048 & 0.035 & 0.026 \\*
  & FSS~$\geq$10 & 0.662 & 0.530 & 0.449 & 0.382 & 0.327 & 0.281 \\*
  & FSS~$\geq$20 & 0.572 & 0.413 & 0.324 & 0.253 & 0.202 & 0.163 \\*
  & FSS~$\geq$40 & 0.437 & 0.279 & 0.200 & 0.140 & 0.103 & 0.078 \\
\midrule
 \multirow{8}{*}{SR $N{=}10$} & RMSE & 0.992 & 0.937 & 1.156 & 1.095 & 1.093 & 1.158 \\*
  & CRPS & 0.0454 & 0.0565 & 0.0654 & 0.0709 & 0.0752 & 0.0795 \\*
  & CSI~$\geq$10 & 0.431 & 0.357 & 0.300 & 0.263 & 0.233 & 0.204 \\*
  & CSI~$\geq$20 & 0.358 & 0.277 & 0.216 & 0.173 & 0.136 & 0.108 \\*
  & CSI~$\geq$40 & 0.263 & 0.185 & 0.132 & 0.089 & 0.058 & 0.038 \\*
  & FSS~$\geq$10 & 0.712 & 0.632 & 0.557 & 0.509 & 0.468 & 0.424 \\*
  & FSS~$\geq$20 & 0.657 & 0.557 & 0.463 & 0.392 & 0.325 & 0.268 \\*
  & FSS~$\geq$40 & 0.561 & 0.442 & 0.341 & 0.245 & 0.166 & 0.113 \\
\midrule
 \multirow{8}{*}{SR $N{=}20$} & RMSE & 0.890 & 0.882 & 1.045 & 1.025 & 1.038 & 1.087 \\*
  & CRPS & 0.0427 & 0.0535 & 0.0619 & 0.0674 & 0.0718 & 0.0759 \\*
  & CSI~$\geq$10 & 0.460 & 0.382 & 0.322 & 0.281 & 0.245 & 0.212 \\*
  & CSI~$\geq$20 & 0.389 & 0.298 & 0.233 & 0.180 & 0.134 & 0.101 \\*
  & CSI~$\geq$40 & 0.296 & 0.197 & 0.139 & 0.082 & 0.045 & 0.025 \\*
  & FSS~$\geq$10 & 0.732 & 0.652 & 0.577 & 0.526 & 0.477 & 0.430 \\*
  & FSS~$\geq$20 & 0.682 & 0.576 & 0.481 & 0.397 & 0.314 & 0.248 \\*
  & FSS~$\geq$40 & 0.596 & 0.454 & 0.346 & 0.222 & 0.129 & 0.074 \\
\midrule
 \multirow{8}{*}{SR $N{=}30$} & RMSE & 0.851 & 0.862 & 1.006 & 1.000 & 1.020 & 1.061 \\*
  & CRPS & 0.0417 & 0.0525 & 0.0608 & 0.0662 & 0.0707 & 0.0747 \\*
  & CSI~$\geq$10 & 0.470 & 0.391 & 0.330 & 0.287 & 0.247 & 0.214 \\*
  & CSI~$\geq$20 & 0.400 & 0.305 & 0.238 & 0.182 & 0.132 & 0.097 \\*
  & CSI~$\geq$40 & 0.307 & 0.200 & 0.139 & 0.077 & 0.040 & 0.021 \\*
  & FSS~$\geq$10 & 0.737 & 0.656 & 0.581 & 0.529 & 0.477 & 0.428 \\*
  & FSS~$\geq$20 & 0.689 & 0.580 & 0.484 & 0.395 & 0.305 & 0.235 \\*
  & FSS~$\geq$40 & 0.606 & 0.454 & 0.341 & 0.208 & 0.112 & 0.060 \\
\end{longtable}

\normalsize
\setlength{\tabcolsep}{6pt}
\renewcommand{\arraystretch}{1.0}

\clearpage


\begin{table}[!htbp]
\small
\caption{\textbf{The correction as a fraction of the backbone rainfall, across lead time (KMA 4\,km grid, full 2023 test period, $N=30$).} The difference between the 30-member SR mean $m_i$ and the exPreCast forecast $b_i$, summed over all pixels $i$ in the evaluation set $\mathcal{M}$ and over all forecasts at the given lead time, and expressed as a fraction of the backbone rainfall over the same pixels: Net $=\sum_{i\in\mathcal{M}}(m_i-b_i)\,/\sum_{i\in\mathcal{M}}b_i$; Absolute $=\sum_{i\in\mathcal{M}}|m_i-b_i|\,/\sum_{i\in\mathcal{M}}b_i$; Ratio $=$ Absolute$\,/\,|$Net$|$. The net term retains the sign of the local differences, so increases and decreases cancel; the absolute term does not, so it accumulates their magnitudes and is the normalized $L^1$ difference between the two fields. Their ratio measures how far the local changes exceed the net change. Results are given for two evaluation sets: the full domain, and the observed rain area $\mathcal{M} = \{i : \mathrm{GT}_i \geq 0.1\,\mathrm{mm\,h}^{-1}\}$. The absolute term reaches $40$--$50\%$, meaning the model locally adds and removes a substantial fraction of the rainfall at each location, while the net term stays at a few percent. Local increases and decreases therefore cancel almost entirely, and precipitation is redistributed in space. The two evaluation sets give the same answer, so the result does not depend on where the comparison is made.}\label{suptab:residual-fraction}
\begin{tabular*}{\textwidth}{@{\extracolsep\fill}lccc}
\toprule
Lead time & Net (\%) & Absolute (\%) & Ratio \\
\midrule
\multicolumn{4}{l}{\textit{Full domain}} \\
$+10$ & $+5.84$ & $33.45$ & $5.7$ \\
$+20$ & $+0.55$ & $35.85$ & $65.3$ \\
$+30$ & $+4.58$ & $39.75$ & $8.7$ \\
$+40$ & $+1.90$ & $40.81$ & $21.5$ \\
$+50$ & $+8.63$ & $43.33$ & $5.0$ \\
$+60$ & $+2.42$ & $45.25$ & $18.7$ \\
Overall & $+3.90$ & $39.53$ & $10.1$ \\
\midrule
\multicolumn{4}{l}{\textit{Observed rain area} ($\text{GT}\geq 0.1\,\text{mm\,h}^{-1}$)} \\
$+10$ & $+6.00$ & $39.65$ & $6.6$ \\
$+20$ & $-1.25$ & $41.74$ & $33.3$ \\
$+30$ & $+3.13$ & $45.82$ & $14.6$ \\
$+40$ & $-0.69$ & $46.56$ & $67.9$ \\
$+50$ & $+7.42$ & $50.46$ & $6.8$ \\
$+60$ & $-1.96$ & $50.83$ & $26.0$ \\
Overall & $+2.05$ & $45.49$ & $22.2$ \\
\bottomrule
\end{tabular*}
\footnotetext{Full 2023 test set (52,325 samples). The SR mean is mean-pooled to the
4\,km grid and the native backbone forecast subtracted from it, as in
Fig. 4 of the main text. The ``Overall'' rows are computed from full-period totals, not as
the arithmetic mean of the per-lead entries, so their ratios are not the mean of the
per-lead ratios.}
\end{table}

\clearpage


\scriptsize
\setlength{\tabcolsep}{4pt}
\renewcommand{\arraystretch}{0.95}
\begin{longtable}{@{}lccccccc@{}}
\caption{\textbf{Full-period conditional verification (KMA 2023 test set).}
Five conditional rates of the 30-member SR Ensemble (1\,km SR output, $P\geq 0.3$ alarm criterion) relative to the exPreCast baseline, evaluated on the 4\,km verification grid across the full 2023 test period (52,325 forecasts at 10-min cadence). Conditional probabilities are defined as in the Methods, writing $\text{Base}$ for the exPreCast forecast field. Event-level values for two high-impact 48-hour windows are reported in main-text Table 2 of the main text. The full-period recovery rates are lower than the event-level values because the test set is dominated by non-precipitation regimes where the baseline already provides near-zero detection, leaving little to recover; the event-level evaluation is therefore the operationally relevant measure for high-impact periods. The false-alarm removal rate exceeds the hit-loss rate at every lead time and threshold, confirming that the correction deletes far more false baseline detections than true ones; the gap narrows in relative terms at the highest threshold and longest lead, where heavy-rain pixels become sparse
}\label{suptab:full-period-recovery} \\
\toprule
Threshold & Metric & \multicolumn{6}{c}{Lead time (min)} \\
\cmidrule(lr){3-8}
(mm\,h$^{-1}$) & & 10 & 20 & 30 & 40 & 50 & 60 \\
\midrule
\endfirsthead
\multicolumn{8}{l}{\footnotesize\textit{Supplementary Table~\ref{suptab:full-period-recovery} (continued)}}\\
\toprule
Threshold & Metric & \multicolumn{6}{c}{Lead time (min)} \\
\cmidrule(lr){3-8}
(mm\,h$^{-1}$) & & 10 & 20 & 30 & 40 & 50 & 60 \\
\midrule
\endhead
\midrule \multicolumn{8}{r}{\footnotesize\textit{Continued on next page}}\\
\endfoot
\botrule
\endlastfoot
 \multirow{5}{*}{$\geq 10$} & Recovery rate (\%)            & 73.98 & 58.11 & 52.50 & 43.66 & 37.43 & 30.70 \\*
  & New false-alarm rate (\%)     & 0.240 & 0.220 & 0.250 & 0.230 & 0.190 & 0.170 \\*
  & Hit-retention rate (\%)       & 99.43 & 98.24 & 97.58 & 95.51 & 92.53 & 90.59 \\*
  & Hit-loss rate (\%)            & 0.57 & 1.76 & 2.42 & 4.49 & 7.47 & 9.41 \\*
  & False-alarm removal rate (\%) & 15.92 & 22.03 & 22.63 & 28.14 & 35.77 & 37.49 \\
\midrule
 \multirow{5}{*}{$\geq 20$} & Recovery rate (\%)            & 69.59 & 49.12 & 43.59 & 31.76 & 24.02 & 17.76 \\*
  & New false-alarm rate (\%)     & 0.100 & 0.080 & 0.100 & 0.080 & 0.060 & 0.050 \\*
  & Hit-retention rate (\%)       & 98.69 & 95.39 & 93.67 & 87.48 & 79.73 & 75.28 \\*
  & Hit-loss rate (\%)            & 1.31 & 4.61 & 6.33 & 12.52 & 20.27 & 24.72 \\*
  & False-alarm removal rate (\%) & 24.51 & 35.71 & 37.36 & 47.98 & 57.30 & 60.29 \\
\midrule
 \multirow{5}{*}{$\geq 40$} & Recovery rate (\%)            & 65.03 & 37.71 & 31.52 & 17.17 & 10.47 & 6.02 \\*
  & New false-alarm rate (\%)     & 0.040 & 0.020 & 0.030 & 0.010 & 0.010 & 0.010 \\*
  & Hit-retention rate (\%)       & 97.03 & 87.74 & 83.30 & 67.80 & 55.54 & 47.58 \\*
  & Hit-loss rate (\%)            & 2.97 & 12.26 & 16.70 & 32.20 & 44.46 & 52.42 \\*
  & False-alarm removal rate (\%) & 35.58 & 54.88 & 57.31 & 72.42 & 80.00 & 83.43 \\
\end{longtable}
\normalsize
\setlength{\tabcolsep}{6pt}
\renewcommand{\arraystretch}{1.0}


\scriptsize
\setlength{\tabcolsep}{4pt}
\renewcommand{\arraystretch}{0.9}
\begin{longtable}{@{}llcccccc@{}}
\caption{\textbf{MeteoNet test set, 1\,km verification grid.} Deterministic and threshold-based verification at $+10$--$60$\,min lead time, evaluated on the native 1\,km grid (no pooling, since the backbone is already at 1\,km) over $2{,}000$ sequences sampled from the test set in proportion to their peak-intensity distribution (the Methods). FSS is computed with a $3\times 3$ neighborhood window; CSI is computed pixel-wise. Thresholds are in dBZ. The SR Ensemble uses the $P\geq 0.3$ alarm criterion. $^{\ast}$For the deterministic baseline (exPreCast), CRPS reduces by definition to MAE; the same value is reported in the CRPS row for direct comparability with the SR Ensemble CRPS. MeteoNet has a native 5-min cadence; values at 10-min steps are shown here for comparability with the KMA tables (Supplementary Tables~\ref{suptab:kma-4km},~\ref{suptab:kma-1km}), and correspond to the curves in main-text Fig. 7a of the main text.}\label{suptab:meteonet} \\
\toprule
Model & Metric & \multicolumn{6}{c}{Lead time (min)} \\
\cmidrule(lr){3-8}
 & & 10 & 20 & 30 & 40 & 50 & 60 \\
\midrule
\endfirsthead

\multicolumn{8}{l}{\footnotesize\textit{Supplementary Table~\ref{suptab:meteonet} (continued)}}\\
\toprule
Model & Metric & \multicolumn{6}{c}{Lead time (min)} \\
\cmidrule(lr){3-8}
 & & 10 & 20 & 30 & 40 & 50 & 60 \\
\midrule
\endhead

\midrule \multicolumn{8}{r}{\footnotesize\textit{Continued on next page}}\\
\endfoot

\botrule
\endlastfoot
 \multirow{8}{*}{exPreCast} & RMSE & 2.130 & 2.626 & 3.022 & 3.333 & 3.599 & 3.832 \\*
  & CRPS & 0.5714 & 0.7117 & 0.8296 & 0.9319 & 1.0184 & 1.0983 \\*
  & CSI~$\geq$28 & 0.562 & 0.467 & 0.402 & 0.354 & 0.317 & 0.287 \\*
  & CSI~$\geq$35 & 0.404 & 0.295 & 0.227 & 0.184 & 0.156 & 0.132 \\*
  & CSI~$\geq$40 & 0.317 & 0.203 & 0.135 & 0.095 & 0.067 & 0.051 \\*
  & FSS~$\geq$28 & 0.854 & 0.763 & 0.691 & 0.632 & 0.583 & 0.540 \\*
  & FSS~$\geq$35 & 0.746 & 0.602 & 0.496 & 0.418 & 0.363 & 0.313 \\*
  & FSS~$\geq$40 & 0.675 & 0.488 & 0.353 & 0.258 & 0.190 & 0.146 \\
\midrule
 \multirow{8}{*}{SR $N{=}1$} & RMSE & 2.574 & 3.164 & 3.593 & 3.925 & 4.205 & 4.440 \\*
  & CRPS & 0.7178 & 0.8853 & 1.0152 & 1.1235 & 1.2051 & 1.2890 \\*
  & CSI~$\geq$28 & 0.480 & 0.388 & 0.329 & 0.287 & 0.254 & 0.226 \\*
  & CSI~$\geq$35 & 0.326 & 0.232 & 0.180 & 0.145 & 0.120 & 0.104 \\*
  & CSI~$\geq$40 & 0.235 & 0.141 & 0.094 & 0.067 & 0.049 & 0.041 \\*
  & FSS~$\geq$28 & 0.799 & 0.698 & 0.618 & 0.559 & 0.506 & 0.463 \\*
  & FSS~$\geq$35 & 0.675 & 0.527 & 0.426 & 0.357 & 0.301 & 0.264 \\*
  & FSS~$\geq$40 & 0.577 & 0.389 & 0.274 & 0.202 & 0.150 & 0.124 \\
\midrule
 \multirow{8}{*}{SR $N{=}10$} & RMSE & 1.990 & 2.478 & 2.834 & 3.105 & 3.328 & 3.511 \\*
  & CRPS & 0.3927 & 0.4966 & 0.5753 & 0.6429 & 0.6923 & 0.7424 \\*
  & CSI~$\geq$28 & 0.540 & 0.460 & 0.407 & 0.369 & 0.338 & 0.312 \\*
  & CSI~$\geq$35 & 0.429 & 0.330 & 0.270 & 0.226 & 0.194 & 0.170 \\*
  & CSI~$\geq$40 & 0.327 & 0.223 & 0.156 & 0.111 & 0.081 & 0.063 \\*
  & FSS~$\geq$28 & 0.811 & 0.736 & 0.680 & 0.636 & 0.598 & 0.566 \\*
  & FSS~$\geq$35 & 0.752 & 0.636 & 0.552 & 0.485 & 0.430 & 0.388 \\*
  & FSS~$\geq$40 & 0.664 & 0.515 & 0.395 & 0.301 & 0.227 & 0.182 \\
\midrule
 \multirow{8}{*}{SR $N{=}20$} & RMSE & 1.953 & 2.434 & 2.785 & 3.054 & 3.273 & 3.453 \\*
  & CRPS & 0.3746 & 0.4749 & 0.5507 & 0.6164 & 0.6640 & 0.7122 \\*
  & CSI~$\geq$28 & 0.560 & 0.478 & 0.425 & 0.385 & 0.352 & 0.326 \\*
  & CSI~$\geq$35 & 0.451 & 0.347 & 0.281 & 0.232 & 0.197 & 0.172 \\*
  & CSI~$\geq$40 & 0.352 & 0.238 & 0.162 & 0.108 & 0.075 & 0.057 \\*
  & FSS~$\geq$28 & 0.820 & 0.745 & 0.689 & 0.645 & 0.606 & 0.575 \\*
  & FSS~$\geq$35 & 0.763 & 0.645 & 0.556 & 0.482 & 0.425 & 0.382 \\*
  & FSS~$\geq$40 & 0.681 & 0.526 & 0.395 & 0.283 & 0.204 & 0.159 \\
\midrule
 \multirow{8}{*}{SR $N{=}30$} & RMSE & 1.940 & 2.419 & 2.769 & 3.036 & 3.254 & 3.434 \\*
  & CRPS & 0.3686 & 0.4677 & 0.5426 & 0.6075 & 0.6546 & 0.7023 \\*
  & CSI~$\geq$28 & 0.567 & 0.485 & 0.431 & 0.390 & 0.357 & 0.331 \\*
  & CSI~$\geq$35 & 0.459 & 0.352 & 0.285 & 0.234 & 0.196 & 0.170 \\*
  & CSI~$\geq$40 & 0.360 & 0.242 & 0.164 & 0.106 & 0.071 & 0.052 \\*
  & FSS~$\geq$28 & 0.823 & 0.748 & 0.691 & 0.646 & 0.608 & 0.576 \\*
  & FSS~$\geq$35 & 0.765 & 0.645 & 0.555 & 0.480 & 0.419 & 0.375 \\*
  & FSS~$\geq$40 & 0.684 & 0.527 & 0.392 & 0.275 & 0.193 & 0.146 \\
\end{longtable}

\normalsize
\setlength{\tabcolsep}{6pt}
\renewcommand{\arraystretch}{1.0}

\begin{figure}[!htbp]
\centering
\includegraphics[width=\textwidth]{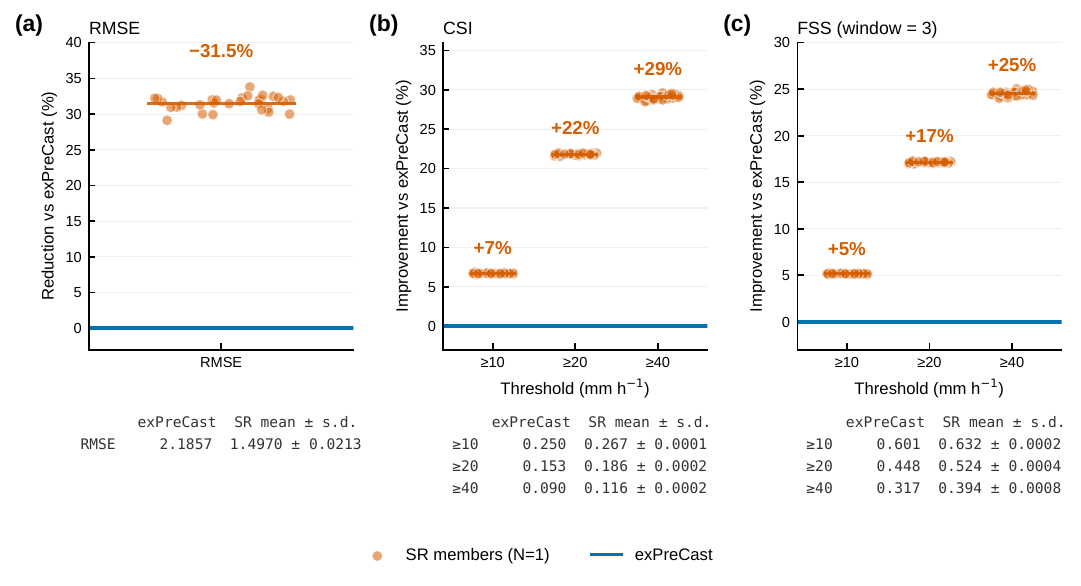}
\caption{\textbf{Seed variability of the single-member forecast (KMA, 4\,km, $N=1$).}
Each of the 30 diffusion members evaluated individually against the exPreCast baseline
over the full 2023 test period, on the 4\,km verification grid. Thresholds, pooling, and
the FSS neighborhood follow Fig. 2a of the main text; scores here are additionally pooled
across lead times, with contingency counts and squared errors accumulated over all
forecasts and lead times before the score is formed.
(a)~RMSE reduction; (b)~CSI improvement and (c)~FSS improvement at the $10$, $20$, and
$40\,\text{mm\,h}^{-1}$ thresholds. Orange points are individual members and the blue line
the baseline; the tables give the baseline value and the member mean $\pm$ standard
deviation. Every member improves on the baseline for all four metrics. The spread across
members is small relative to the margin over the baseline --- $0.0213$ against a reduction
of $0.69$ in RMSE, and at most $0.0008$ against improvements of $0.017$--$0.033$ in CSI
--- so the $N=1$ results reported elsewhere do not depend on the particular member drawn.
Lead-time-resolved values for the nested ensembles are given in Supplementary
Table~\ref{suptab:kma-4km}.}
\label{supfig:seed-robustness}
\end{figure}
\clearpage
\bibliography{sn-bibliography}
\end{document}